\documentclass[mnsc,nonblindrev]{informs3} 

\OneAndAHalfSpacedXI 
\usepackage[hidelinks]{hyperref}

\usepackage{endnotes}
\let\footnote=\endnote
\let\enotesize=\normalsize
\def\notesname{Endnotes}%
\def\makeenmark{$^{\theenmark}$}
\def\enoteformat{\rightskip0pt\leftskip0pt\parindent=1.75em
  \leavevmode\llap{\theenmark.\enskip}}

\usepackage{natbib}
 \bibpunct[, ]{(}{)}{,}{a}{}{,}%
 \def\bibfont{\small}%

\usepackage{algorithm}
\usepackage{algorithmicx}
\usepackage{algpseudocode}
\algnewcommand{\algorithmicand}{\textbf{and }}
\algnewcommand{\algorithmicor}{\textbf{or }}
\algnewcommand{\OR}{\algorithmicor}
\algnewcommand{\AND}{\algorithmicand}
\makeatletter
\newenvironment{breakablealgorithm}
  {
  \begin{center}
     \refstepcounter{algorithm}
     \hrule height.8pt depth0pt \kern2pt
     \renewcommand{\caption}[2][\relax]{
      {\raggedright\textbf{\ALG@name~\thealgorithm} ##2\par}%
      \ifx\relax##1\relax 
         \addcontentsline{loa}{algorithm}{\protect\numberline{\thealgorithm}##2}%
      \else 
         \addcontentsline{loa}{algorithm}{\protect\numberline{\thealgorithm}##1}%
      \fi
      \kern2pt\hrule\kern2pt
     }
  }{
     \kern2pt\hrule\relax
  \end{center}
  }
\usepackage{makecell}

\usepackage{ulem}
\usepackage{booktabs}
\usepackage{multirow}
\usepackage{comment}

\TheoremsNumberedThrough     
\ECRepeatTheorems

\EquationsNumberedThrough    

\newenvironment{assumption'}[1]
  {\renewcommand{\theassumption}{\arabic{assumption}$'$}%
   \addtocounter{assumption}{-1}%
   \begin{assumption}}
  {\end{assumption}}

\newcommand{\UB}{\textbf{\rm UB}}
\newcommand{\val}{\textbf{\rm val}}
\newcommand{\cO}{\mathcal O}
\newcommand{\cR}{\mathcal R}
\newcommand{\cD}{\mathcal D}
\newcommand{\cI}{\mathcal I}
\newcommand{\cc}{\text{\rm cc}}

\usepackage{amsmath}
\usepackage{amsfonts}
\usepackage{amssymb}
\usepackage{dsfont}
\newenvironment{alg'}[1]
  {\renewcommand{\thealgorithm}{\ref{#1}$'$}%
   \addtocounter{algorithm}{-1}%
   \begin{breakablealgorithm}}
  {\end{breakablealgorithm}}

\newenvironment{lemma'}[1]
  {\renewcommand{\thelemma}{\ref{#1}$'$}%
   \addtocounter{lemma}{-1}%
   \begin{lemma}}
  {\end{lemma}}

\begin{document}



\RUNTITLE{Offline Planning and Online Learning under Recovering Rewards}

\TITLE{Offline Planning and Online Learning under Recovering Rewards}


\ARTICLEAUTHORS{%
\AUTHOR{David Simchi-Levi}
\AFF{Institute for Data, Systems, and Society, Massachusetts Institute of Technology, Cambridge, MA 02139}
\AUTHOR{Zeyu Zheng}
\AFF{Department of Industrial Engineering and Operations Research, University of California, Berkeley, CA 94709}
\AUTHOR{Feng Zhu}
\AFF{Institute for Data, Systems, and Society, Massachusetts Institute of Technology, Cambridge, MA 02139}
} 

\ABSTRACT{%
Motivated by emerging applications such as live-streaming e-commerce, promotions and recommendations, we introduce and solve a general class of non-stationary multi-armed bandit problems that have the following two features: (i) the decision maker can pull and collect rewards from up to $K\,(\ge 1)$ out of $N$ different arms in each time period; (ii) the expected reward of an arm immediately drops after it is pulled, and then non-parametrically recovers as the arm's idle time increases. With the objective of maximizing the expected cumulative reward over $T$ time periods, we design a class of ``Purely Periodic Policies'' that jointly set a period to pull each arm. For the proposed policies, we prove performance guarantees for both the offline problem and the online problems. For the offline problem when all model parameters are known, the proposed periodic policy obtains an approximation ratio that is at the order of $1-\cO(1/\sqrt{K})$, which is asymptotically optimal when $K$ grows to infinity. For the online problem when the model parameters are unknown and need to be dynamically learned, we integrate the offline periodic policy with the upper confidence bound procedure to construct on online policy. The proposed online policy is proved to approximately have $\widetilde\cO(N\sqrt{T})$ regret against the offline benchmark. Our framework and policy design may shed light on broader offline planning and online learning applications with non-stationary and recovering rewards.
}%


\KEYWORDS{offline planning, online learning, recovering rewards} 
%

\maketitle

%


\section{Introduction}

In the past decades, the study of sequential decision making has attracted prominent attention from academic researchers and industrial practitioners. Emerging applications such as live-streaming e-commerce, promotions, and recommendations have given rise to new features and posed challenges to classical sequential decision making tools. In this paper, using the framework of multi-armed bandits problems, we introduce and solve a general class of sequential decision making problems that accommodate two relevant new features. The first feature is that, in each time period, the decision maker can simultaneously pull and collect rewards from at most $K$ arms out of a total of $N$ arms. When $K$ is an arbitrary number that is greater than one, this feature generalizes the standard multi-armed bandits problems, accommodating applications where the decision maker can pull more than one arm simultaneously in one time period. The second feature is that, the expected reward of an arm immediately drops after it is pulled, and then gradually recovers if the arm is not pulled in the subsequent time periods. This feature matches a number of phenomena in the area of management science, including one that customers may temporarily get bored if they are repeatedly recommended with the same product, same deal, or same type of promotion. The general class of multi-armed bandits problems with these two features may appear in broad applications that involve non-stationary and dynamically changing rewards. 

We use the example of live-streaming e-commerce to illustrate the problem setting and two new features, in a simplified description. The rapidly growing live-streaming e-commerce is estimated to hit \$100 billion in annual global sales in the year of 2020 and is expected to exceed \$300 billion in the year of 2021; see \citet{Greenwald2020ecommerce}, \citet{Kharif2020e} and \citet{Enberg2021ecommerce}. A popular form of live-streaming e-commerce is session-based promotions held by a key opinion leader (KOL) on platforms such as Amazon Live, Taobao Live and TikTok. In session-based promotions, a KOL holds sessions on consecutive days. 
For each session, the KOL chooses say about $K=50$ products to discuss their features, promote and sell during the session, one product after another, often at deep discounts. The prepared inventory for the promoted products can often sell out in seconds. Many manufacturers pitch top KOLs and create a large pool of products for a KOL to choose from, with an estimated ten to one selection ratio \cite{Greenwald2020ecommerce, Kharif2020e}. Therefore, the entire pool of products can be of size $N=500$. This emerging application of live-streaming e-commerce gives rise to the need for dynamic planning and learning in presence of the two aforementioned features. First, in each time period, the decision maker needs to select $K=50$ products (arms) to promote and sell out of a pool of $N=500$. Second, if a product is promoted and on sale at some time period $t$ with expected reward $R$, the expected reward of promoting and selling the same product again in time period $t+1$ will drop and can be much smaller than $R$. On the other hand, as the idle time (number of time periods that a product has not been promoted) increases, the expected reward of the product can increase. See, as an example, Section 4 of \cite{shen2020JD} for empirical observations that may support the feature that the daily sales/revenue drop after a strong promotion day or period. Because of this feature, even for the most popular product, a KOL will not promote and sell the product in every single session, but alternatively, tend to select the product in every few other sessions. 

Aside from the example of live-streaming e-commerce, there are a number of other applications that exhibit such rewards dropping and recovering phenomenon. For example, in the recommendation domain, a customer may enjoy a particular product or service, but may temporarily get bored immediately after consuming the product or service. In this case, the reward of recommending the same product or service to the customer will temporarily drop after the most recent consumption, but may gradually recover as time passes by. Different products or services can have different dropping and recovering behavior, and it is of the decision maker's interest to strategically choose when and how often to recommend each product or service. For any given product, a resting time between each time it is recommended can be valuable, which intuitively offers a refreshing window for the customer. 

With the incorporation of the two new features, solving the proposed general class of sequential decision-making problems present new challenges and difficulties. The difficulties are centered around algorithm design and the establishment of theoretical guarantees. Specifically, even for the offline problem where all model parameters are initially known but future realizations of uncertainties are unknown, it is computationally hard to obtain a near-optimal policy. First, in practice, the reward recovery mechanism typically may not exhibit nice structural patterns such as concavity or parametric forms. In presence of the potentially complicated reward recovery mechanism, it is challenging to obtain simple-to-implement policies that fit to a broad range of application settings with theoretical performance guarantees. Second, the reward for pulling an arm is concerned with not only the current time period, but also the time period when the arm was pulled last time, which is essentially affected by the policy itself. Thus, unlike classical bandits problems, in this problem there does not exist an ``optimal'' subset of arms to identify and eventually pull for each time period. The decision maker has to adjust the subset of pulled arms dynamically in a delicate way.

For the online problem where parameters need to be learnt on-the-fly, we also face difficulties that are distinguishable from canonical online learning problems. First, apart from the classical exploration-exploitation dilemma, we also face with a planning-learning trade-off. To improve the knowledge of the reward recovery mechanism, we need to plan for a delay to collect samples. However, planning for a long delay may cause a waste of time and insufficient learning. Second, the design of the offline optimization oracle in this online setting needs additional treatment. Typically, offline oracles are assumed to be powerful enough to solve given inputs with theoretical guarantees. However, due to the estimation errors caused by randomness, it is unclear whether the offline planning result can be directly adapted into the online learning one.

In this paper, we overcome the difficulties mentioned above. We summarize our contributions as follows.

\begin{enumerate}
\item \textbf{A General Recovering Model}: To characterize the dropping and recovering behavior of the rewards which is represented by a recovering function, we build a non-parametric bandit model with minimal assumptions on the recovery functions: monotonicity and boundedness. The recovery functions do not need to satisfy \textit{any other} structural properties such as concavity. Our model relaxes assumptions and generalizes settings in previous work, thus allowing wider potential applications.

\item \textbf{An Offline Planning Framework}: Even if the recovery functions are fully known, the planning problem of joint optimally choosing $K$ out of $N$ arms for each of the $T$ time periods is computationally difficult. We construct an upper bound on the offline planning problem, which leads us to focus on a simple and intuitive class of policies called ``Purely Periodic Policies'' (PPP). Surprisingly for the proposed policies, the long-run offline approximation ratio (the ratio between the expected cumulative reward of the constructed policy and the optimal) only depends on $K$ (which we denote as $\gamma_K$) and does not depend on any structure of the recovery function. This ratio, moreover, has a uniform lower bound $1/2$ and approaches $1$ with a $\cO(1/\sqrt{K})$ gap uniformly over $K\geq 1$. Further, we prove that this $\cO(1/\sqrt{K})$ gap is tight within the class of PPP up to a logarithm factor, showing that any other PPP cannot do better than our algorithm asymptotically. The framework of our policy design and analysis is novel, in the sense that we utilize power of $2$ to bypass the difficulty of combinatorial planning, and address the trade-off between rounding error and scheduling error. Moreover, our policy design framework does not rely on standard parametric and structural assumptions and can potentially be adapted to various other applications.

\item \textbf{A Learning-while-Planning Algorithm}: We address the problem where we have no prior knowledge of the recovery functions, and must learn through online samples. Enlightened by the offline planning part, we propose an online learning policy to solve the exploration-exploitation trade-off, and moreover, reach a balance between planning and learning. The policy incurs a $\widetilde\cO(\sqrt T)$ regret compared to our offline result. Our design of the online learning algorithm exploits the sparse structure of the ``Purely Periodic Policy'' proposed in our offline planning framework, relates it to a generalization of the knapsack problem, and integrates it with upper confidence bounds, which may bring new insights to solve other online learning problems under recovering rewards.
\end{enumerate}

\subsection{Related Literature}

\textbf{Multi-armed Bandits:} The Multi-armed Bandit (MAB) problem has aroused great interest in both academia and industry in the last decade. Since the pioneer work of \citet{auer2002finite}, researchers have developed many theoretically guaranteed algorithms based on the ``optimism under uncertainty'' principle. Readers could refer to \citet{lattimore2019bandit} and \citet{slivkins2019introduction} for a comprehensive discussion about this field. An important MAB variant is the combinatorial MAB problem, where in each time the player can pull a subset of arms instead of only one arm (see, e.g., \citealt{kveton2015tight}). This better captures real-world applications, and we also adopt the setting.

Another variant of MAB arousing interest in recent years is non-stationary MAB, where the mean reward of each arm can change with time. Two main non-stationary models are restless bandits \cite{whittle1988restless} and rested bandits \cite{gittins1979bandit}. In restless bandits, the mean reward of an arm changes irrespective of the policy being used, while in rested bandits, the mean reward changes only when the arm is pulled by the policy. These two problems have been investigated through many different perspective, such as variation budget \cite{besbes2014stochastic}, abrupt changes \cite{auer2019adaptively}, rotting bandits \cite{levine2017rotting}, and applications in dynamic pricing \cite{keskin2017chasing, zhu2020demands}. We note that our setting is neither rested nor restless, as our reward distribution of an arm changes according to both the time period itself and whether the arm is selected by the policy or not. Our problem exhibits a recovering behavior: the mean reward of an arm first drops after a pulling, but will gradually recover as the idle time increases. Below we review the bandits literature with recovering rewards, which is most related to our work.

\textbf{Bandits with Recovering Rewards:} Recently there has been an increasing number of work studying bandits under different recovering environments. However, so far all the work only consider pulling one arm at a time. The work most relevant to ours is \citet{kleinberg2018recharging}. The authors develop a PTAS to obtain a long-run $(1-\epsilon)$-optimal scheduling for the offline problem of which the time complexity is doubly exponential in $1/\epsilon$, and derive a $\widetilde\cO\left(\frac{\sqrt{T}}{\epsilon^{\cO(1/\epsilon)}}\right)$ regret for the online problem by adapting the offline result. Their analysis is heavily dependent on the assumption that the recovery function is non-decreasing and globally concave. We remove this somewhat restrictive assumption by only assuming the monotonicity, allowing more general behavior after a pull. In this case, the property of the randomized policy discussed in \citet{kleinberg2018recharging} may no longer be valid, and thus we propose different solution methods and new rounding techniques.

A special case of our model is investigated in \citet{basu2019blocking}, where an arm cannot be pulled for a while after it is pulled. The ``sleeping time'' of each arm is known a priori. The authors compete their online learning algorithm with the greedy policy which yields a long-run $1-1/e$ approximation guarantee. Because of the nice property of the greedy policy, some subsequent work follows this line and considered more general blocking bandit models (see, e.g., \citealt{basu2020contextual}, \citealt{atsidakou2021combinatorial}). However, in our more general setting, the greedy policy may perform very bad compared to the true optimum. Also, in our online learning problem, we have no prior knowledge of the sleeping times. \citet{cella2020stochastic} generalizes the setting in \citet{basu2019blocking} using non-parametric recovery. However, they assume that the recovery rate is the same among all arms, and only consider purely periodic policies where all selected arms are pulled at the same frequency with no constant ratio guarantees. We adopt a more generalized model and take a broader approach by allowing different recovery rates and different pulling frequencies for different arms, and yields a $1/4$ worst-case guarantee for the offline problem.

Another related work is \citet{pike2019recovering}, where the expected reward functions are sampled from a Gaussian Process with known kernel. The authors compare their algorithm to a Bayesian $d$-step lookahead benchmark, which is the greedy algorithm optimizing the next $d$ pulls given the decision maker’s current situation. In comparison, our benchmark is concerned with the total reward of the whole time horizon $T$ rather than a pre-fixed $d$. Some other related work include \citet{mintz2020nonstationary} and \citet{yancey2020sleeping}. In \citet{mintz2020nonstationary}, the recovery function is characterized via a parametric form, while the authors obtain a worst-case regret of $\widetilde\cO(T^{\frac{2}{3}})$. \citet{yancey2020sleeping} considers a specific application for scheduling of reminder notifications. The authors demonstrate their policy via numerical experiments. We note that our problem is non-parametric in essence, and for the online problem we have a theoretical guarantee of $\widetilde\cO(\sqrt{T})$ regret compared to our offline benchmark. 

\section{Model and Assumptions} \label{ssec:model}
There are $N$ arms in total. Denote $[N]$ as the set $\{1,2,\ldots,N\}$. Let $K$ be any positive integer that satisfies $K\le N$. In each time period $t=1,2,3,\ldots$, the action is to choose a set of no more than $K$ arms from the total $N$ arms and simultaneously pull the chosen arms. The revenue collected in each time period is the sum of the reward of each pulled arm. The expected reward of an arm is not stationary and may change over time in the following way. For each arm $i$, consider a set of non-negative scalars $\{R_i(d)\}_{d=0,1,2,\ldots}$ that construct the \textit{recovery function} $R_i(\cdot)$ with respect to $d$. The reward of pulling arm $i$ at time period $t$ is a random variable with mean $R_i(t-t')$, where $t'$ denotes the most recent time period (prior to $t$) when arm $i$ is pulled. We let $R_i(0)=0$ for completeness. The randomness in the reward of pulling an arm $i$ at time $t$ is assumed to be independent of all other sources of randomness in the system. The system runs from time period $1$ and continues in discrete time periods $2,3,\ldots$. To clearly define the initialization of the system at time period $1$, one needs to specify when is the last time period that each arm is pulled before time period $1$. There have been two common types of initialization. One is that the last time that each arm is pulled is $-\infty$; see \citet{basu2019blocking}. The other is that the last time that each arm is pulled is time period $0$, right before the start of the problem at time period $1$; see \citet{kleinberg2018recharging}. We adopt the second type of initialization, and we note that there is no intrinsic difference between the first type and second type regarding the analytical results to be established in this work. 

Next we describe and discuss assumptions on the recovery function. 

\begin{assumption}[\textbf{Monotonicity}] \label{assumption:mono}
For any $i\in[N]$, the recovery function $\{R_i(d)\}_{d\geq 0}$ is non-decreasing in $d$. 
\end{assumption}

The non-decreasing property of the expected reward sequence $\{R_i(d)\}_{d\geq 0}$ reflects the modeling feature that the longer the time for which an arm has been idle (not pulled), the larger the expected reward once the arm is pulled again. In fact, the range of reward models implied by Assumption \ref{assumption:mono} includes that implied by the assumptions made in the literature --- concave recharging bandits \cite{kleinberg2018recharging}, sleeping/blocking bandits \cite{basu2019blocking}, and recovering bandits with identical recovery rate \cite{cella2020stochastic}. Specifically, we extend the concave non-decreasing structure in the literature to general non-decreasing structure, which assumes no specific function shape and allows more applications.

\begin{assumption}[\textbf{Boundedness}] \label{assumption:bound}
The reward of pulling an arm in each time period, as a random variable, is mutually independent and is bounded by a known constant $R_{\max}$ uniformly over all $i\in[N]$.
\end{assumption}

The assumption of bounded random rewards is common in many applications such as promotion in live-streaming e-commerce and recommendation. An alternate but slightly more restrictive assumption to Assumption \ref{assumption:bound} for the mean rewards $\{R_i(d)\}_{i\in[N], d\geq 0}$ is called ``Finite-time Recovery''.

\begin{assumption'}{assumption:bound}[\textbf{Finite-time Recovery}] \label{assumption:finite} For any $i\in[N]$, $\exists d_i^{\max}\geq 1$ such that $R_i(d) = R_i(d_i^{\max})$ for all $d\geq d_i^{\max}$.
\end{assumption'}

Assumption \ref{assumption:finite} means that the mean reward of each arm recovers to a max nominal level and stays there after a finite number of idle time periods (see also, e.g., \citealt{basu2019blocking, pike2019recovering}). We will see that this assumption is added in Section \ref{sec:offline} primarily for computational concerns. All of our theoretical results and bounds do not depend on $\{d_i^{\max}\}$ that can be large in practice.

\section{Offline Planning Problem} \label{sec:offline}
In this section, we describe the definition of a decision-making problem that we refer to as the \textit{offline planning problem}. We discuss the challenges of solving this problem and propose a class of easy-to-implement offline policies that can alleviate these challenges. 

\subsection{Problem Statement} \label{ssec:offline-statement}

The offline planning problem is given as follows. The knowledge of $\{R_i(d)\}_{i\in[N], d\geq 0}$ is fully known to the decision maker and there is no unknown parameter. The randomness in the system does not impact the policy design for the offline problem, because all the model parameters are known and there is no need to learn those parameters under uncertainties. The objective is to maximize the total expected reward throughout a time horizon of $T$, under the constraint that no more than $K$ arms can be pulled in each time period. More precisely, we define $\mathrm{OPT}_\cR[K, T]$ (sometimes abbreviated as $\mathrm{OPT}$) as the maximal expected reward that can be obtained among all different policies:
\begin{align}
    \mathrm{OPT}\, \triangleq \quad \max \quad & \sum_{i\in[N]}\sum_{j=1}^{J_{i}}R_i(s_{i, j}-s_{i, j-1}) \label{program:OPT}\\
    \text{s.t.} \quad & J_i\in[T], \quad \forall i\in[N], \nonumber \\
    & 0=s_{i, 0} < s_{i, 1} < \cdots < s_{i, J_i} \leq T, \quad \forall i\in[N], \nonumber\\
    & \sum_{i\in[N]}\sum_{j\in J_i}\mathds 1\{s_{i, j}=t\} \leq K, \quad \forall t\in[T]. \nonumber
\end{align}
Here, $J_i\in[T]$ is the number of times arm $i$ is pulled. $s_{i, 1}, \cdots, s_{i, J_i}$ is the time schedule of pulling arm $i$. The last group of constraints in (\ref{program:OPT}) upper bounds the number of pulled arms in each time period by $K$.

We note that exactly solving the offline planning problem (\ref{program:OPT}) is computationally intractable and is NP-hard. The problem setting in \citet{basu2019blocking} is a special case of ours, and they discuss in Section 3 the computational intractability of their problem setting. Since our problem setting is more general and in addition allows arbitrary non-parametric recovery, the optimal solution can be even more complicated because of the non-parametric recovery. 

Our goal is to propose a framework of policies that enjoy two properties. First, the proposed policies have provable performances.  
Specifically, we hope to find an absolute constant $\gamma_K>0$ possibly dependent on $K$, and another constant $\beta$ that does not depend on $T$, such that the total expected reward collected by the proposed policies is at least
\begin{align*}
    \gamma_K\cdot \mathrm{OPT} + \beta
\end{align*}
for all reward recovery instance $\cR=\{R_i(d)\}_{i\in[N], d\geq 0}$. Here, $\gamma_K$ is the (long-run) approximation ratio, $\mathrm{OPT}$ scales up with $T$ and an additional $\beta$ term that does not depend on $T$ is allowed. This notion of approximation ratio is common in previous work on bandits with an recovering mechanism (see, e.g., \citealt{kleinberg2018recharging}, \citealt{basu2019blocking}).

Second, the proposed policies should be easy to construct and implement. Mathematically, the policies should be run in polynomial time with $K, N, T$.  Moreover, the policies should be flexible enough to be adapted to more complex settings such as the online learning problem where the parameters are unknown a priori. A first thought is to consider the use of greedy policy, which was widely used and was shown to have reliable performances in many settings including bandits problems with recovering rewards (\citealt{kleinberg2018recharging}, \citealt{basu2019blocking}, \citealt{basu2020contextual}).  The greedy policy in our setting means that the policy selects $K$ arms with the highest expected reward in the current time period. However, it fails on the first point regarding the performance guarantee. In fact, the greedy policy may have arbitrarily bad expected total reward in terms of the ratio with the optimal, as is shown in Example \ref{example:greedy:bad}.

\begin{example} \label{example:greedy:bad}
Let $N=2$ and $K=1$. Let $R_1(d) = r$ and $R_2(d) = \mathds 1_{d=1} + R\cdot\mathds 1_{d>1}$, where $R\gg 1 > r$. Then under the greedy policy, we will always pull arm 2, yielding a total reward of $T$. However, if we pull arm 1 and 2 in turn, then the total reward is lower bounded by $\frac{R+r}{2}(T-1)\gg T$ when $R\gg 1$. Therefore, the ratio between the expected total reward achieved by the greedy policy and the optimal can be arbitrarily small.
\end{example}

Note that as a comparison, in the problem settings of \citet{kleinberg2018recharging} and \citet{basu2019blocking}, the greedy policy in an analogous form, despite of being sub-optimal, has a guaranteed lower bound at least $1/2$ in terms of the ratio between the expected total reward achieved by the greedy policy and the optimal. This also suggests that our problem setting that allows arbitrary non-parametric recovery has more subtleties. 


\subsection{Purely Periodic Policies (PPP)}

We have seen above that a naive greedy policy may perform not well for general $\cR$ in the worst case. In the following, we alternatively propose the use of a simple and easy-to-implement class of policies called \textit{purely periodic policies}.

\begin{definition}[\textbf{Purely Periodic Policy}] \label{def:PPP}
A policy is called ``purely periodic'', if for each $i\in[N]$, the policy assigns $t_i\in\mathbb Z$ and $d_i\in\mathbb Z_+\cup\{+\infty\}$ such that (i) $t_i \in (-d_i, 0]$, and (ii) arm $i$ is pulled at time $t_i+k\cdot d_i\ (k\in\mathbb Z_+)$ until the time horizon $T$ is exhausted. A purely periodic policy is represent by $\{(d_i, t_i)\}_{i\in[N]}$.
\end{definition}

We note that for a purely periodic policy, when $d_i=+\infty$, $t_i\leq 0$ for some $i$, the policy never pulls arm $i$ throughout the whole time horizon. In the following, we use ``PPP'' as the brief notation of ``purely periodic policy''.

\begin{definition}[\bf $K$-PPP] \label{def:K-PPP}
Let $K\in\mathbb Z_+$. A policy is called a $K$-PPP, if the policy is a purely periodic policy $\{(d_i, t_i)\}_{i\in[N]}$, and meanwhile in each time period $t\geq 1$, at most $K$ arms are pulled at the same time. Equivalently speaking, 
\begin{align*}
    \sup_{t\geq 1}\#\left\{i\in[N]: d_i < +\infty, t\equiv t_i\ (\bmod\ d_i)\right\} \leq K.
\end{align*}
\end{definition}

With an overall number of $N$ arms, each PPP is always an $N$-PPP. For the general case where $K\leq N$, the set of $K$-PPPs is a subset within the class of PPPs. Despite its simplicity, the family of PPPs indeed owns nice properties. On the computational side, it can be constructed in polynomial time. On the theoretical side, the long-run approximation ratio $\gamma_K$ is lower bounded by an absolute constant and approaches $1$ as $K$ increases. It turns out that these properties are quite non-trivial, and we will devote the next section to show how. We note that finding the long-run optimal policy for the offline problem is NP-Hard even if we confine ourselves within the class of $K$-PPP. In the supplementary material, we give a proof demonstrating this point.

There are three main challenges for constructing a $K$-PPP that delivers superior performance. First, we need to determine the set of arms that are included in the planning schedule: $\{i: d_i<+\infty\}$. Second, we need to find appropriate frequencies $\{1/d_i\}$ for selected arms, or equivalently, the set of periods $\{d_i\}$. Finally, we need to pair each $d_i$ with some $t_i$, as defined in Definition \ref{def:K-PPP}. The $\{d_i\}$'s must be selected carefully, or there may not exist such $\{t_i\}$ that no more than $K$ arms can be pulled at the same time. We illustrate this point through the example below.
\begin{example}
Let $N=2$ and $K=1$. Then there exists no $1$-PPP such that $d_1=2$ and $d_2=3$, since 2 and 3 are co-prime numbers. No matter how we choose $t_1$ and $t_2$, there must exist a time period such that both arms are scheduled to be pulled together, violating the constraint $K=1$. However, if $d_1=2$ and $d_2=4$, then letting $t_1=-1$ and $t_2=-2$ yields a feasible $1$-PPP.
\end{example}

\section{Bounds and Policy Construction for \texorpdfstring{$K$}{}-PPP} \label{sec:PPP}

In this section, we aim at constructing a class of $K$-PPPs that obtain theoretical guarantees through the analysis of an upper bound on $\text{OPT}$ from the offline planning problem. We first discuss how to construct a reasonable upper bound and its implications on PPP. Then, we propose a rounding-scheduling (R-S) framework about constructing a $K$-PPP. Finally, we present (long-run) approximation guarantees for our proposed algorithm as well as its tightness.

\subsection{Bounding the Objective: Why Purely Periodic?} \label{ssec:offline-upper-bound}

The goal of this part is to define a reasonable ``baseline'' for our problem. Specifically, we will give an upper bound on the optimal (expected) objective value of the problem ($\text{OPT}$) when the rewards recovery functions are fully known. Meanwhile, we will give an interpretation on the advantages of PPP via the analysis on the upper bound. We start our discussion from the rewards recovery functions $\{R_i(d)\}_{i\in[N], d\geq 0}$.

For given $i$, let $\{R_i^{\cc}(d)\}_{d\geq 0}$ be the upper concave envelope of $\{R_i(d)\}_{d\geq 0}$. That is, $R_i^{\cc}$ is the point-wise minimum of all concave functions $F: \mathbb Z_+\cup\{0\}\to \mathbb R$ that satisfy $F(d)\geq R_i(d)$ for all non-negative integer $d$. We define \textit{supporting points} as the set of points $\{d\geq 0: R_i(d) = R_i^{\cc}(d)\}$. Then these supporting points can be characterized inductively as follows. $d_i^{(0)}=0$. For any $k\geq 1$, 
\begin{align*}
    d_i^{(k)} = \min\left\{\arg\max_{d>d_i^{(k-1)}}\left\{\frac{R_i(d) - R_i(d_i^{(k-1)})}{d - d_i^{(k-1)}}\right\}\right\}.
\end{align*}
When $\{R_i(d)\}$ is bounded, for any $k\geq 1$, we have
\begin{align*}
\lim_{d\to+\infty} \frac{R_i(d) - R_i(d_i^{(k-1)})}{d - d_i^{(k-1)}} = 0.    
\end{align*}
If
\begin{align*}
\max_{d>d_i^{(k-1)}}\left\{\frac{R_i(d) - R_i(d_i^{(k-1)})}{d - d_i^{(k-1)}}\right\} > 0,
\end{align*}
then $d_i^{k} < +\infty$. Otherwise, $d_i^{k}=d_i^{(k-1)}+1$. In both cases, $\{d_i^{(k)}\}_{k\geq 0}$ is an infinite sequence for any given $i\in[N]$. When Assumption \ref{assumption:finite} holds, the supporting points of the concave envelope for arm $i$ can be efficiently obtained by only examining $\{R_i(d)\}_{0\leq d\leq d_i^{\max}}$.

Let $F_{i, T}(x)$ be the value of the following problem.
\begin{align} \label{program:upper-bound-single}
     \max_s \quad & \frac{1}{T}\sum_{j=1}^J R_i(s_j-s_{j-1}) \\
    \text{s.t.} \quad & J \leq x\cdot T, \nonumber\\
    & 0 = s_0 < s_1 < \cdots < s_J \leq T. \nonumber
\end{align}
Intuitively, $F_{i, T}(x)$ is the optimal average reward of $i$ given that we pull arm $i$ no more than $x\cdot T$ times. Let $F_i(x) = \limsup_{T}F_{i, T}(x)$ be the point-wise long-run optimal average reward of pulling arm $i$ under $x$. We have Lemma \ref{lemma:F-property} that fully characterizes the nice shape of $F_i(x)$.

\begin{lemma} \label{lemma:F-property}
$F_i(x)$ is a piece-wise linear concave function on $[0, 1]$ that satisfies:
\begin{itemize}
    \item The changing points are in $\{1/d_i^{(k)}\}_{k\geq 1}$.
    \item $F_i(1/d_i^{(k)}) = R_i(d_i^{(k)})/d_i^{(k)}$ for any $k\geq 1$. $F_i(x) = F_i(1/d_i^{(1)})$ for all $x\in[1/d_i^{(1)}, 1]$.
\end{itemize}
Meanwhile, $F_i(x)\geq F_{i, T}(x), \forall i\in[N], T\geq 1, x\in[0, 1]$.
\end{lemma}

We are then ready to present the upper bound (\ref{program:upper-bound}). 
\begin{align} \label{program:upper-bound}
    \max_{x\in[0, 1]^N} \quad & T\sum_{i=1}^N F_i(x_i) \\
    \text{s.t.} \quad & \sum_{i=1}^N x_i \leq K. \nonumber
\end{align}

\begin{lemma} \label{lemma:upper-bound}
The optimal objective value of (\ref{program:upper-bound}) is an upper bound on that of the offline problem. Further, problem (\ref{program:upper-bound}) has an optimal solution $\{x_i^*\}$, such that $x_i^*\leq 1/d_i^{(1)}\ (\forall i\in[N])$, and for at least all $i\in[N]$ but one, $x_i^*\in\{0\}\cup\{1/d_i^{(k)}\}_{k\geq 1}$ holds.
\end{lemma}

We give some remarks on Lemma \ref{lemma:F-property} and \ref{lemma:upper-bound}. The two lemmas are generalizations of Lemma III.1 and III.2 in \citet{kleinberg2018recharging} as we do not assume \textit{any} concavity of recovery functions. The proofs of both lemmas only require the existence of \textit{supporting points}. Moreover, we go beyond their results by observing that almost each component of the optimal solution is either 0 or $1/d$. In the proof of Lemma \ref{lemma:upper-bound}, we demonstrate that any feasible solution can be transformed into one that satisfies the property stated in Lemma \ref{lemma:upper-bound} in $\cO(N)$ time, while at the meantime the objective value is not decreased. This crucial property can be interpreted as in the optimal planning, we either pull arm $i$ once every $d$ times, or we do not pull it at all, which is exactly the form of PPP (see Definition \ref{def:PPP}). 
In the following discussion, we denote the objective value of (\ref{program:upper-bound}) as $\text{UB}_{\cR}[K]\cdot T$, where $\cR = \{R_i(d)\}_{i\in[N], d\geq 0}$ is the set of reward recovery scalars for all $N$ arms.

Before ending this section, we give some remarks on the computation of (\ref{program:upper-bound}). Under Assumption \ref{assumption:finite}, an optimal solution of the concave program (\ref{program:upper-bound}) can be efficiently computed because (\ref{program:upper-bound}) can be re-written as a linear program. \citet{kleinberg2018recharging} considered the case where $K=1$ and $d_i^{(k)}=k\ (\forall k\geq 0)$, and suggests that (\ref{program:upper-bound}) admits an 
FTPAS even if Assumption \ref{assumption:finite} is violated. 
It is thus interesting to investigate whether (\ref{program:upper-bound}) admits a polynomial-time algorithm in general and we leave this to future work.

\subsection{Constructing a \texorpdfstring{$K$}{}-PPP} \label{ssec:offline-schedule}

We now discuss how to take \textit{any} $x\in[0, 1]^N$ and outputs a $K$-PPP through two steps. Here $x$ does no need to satisfy the first constraint of (\ref{program:upper-bound}), but the output schedule is always guaranteed to satisfy the hard constraint $K$. We emphasize that in contrast to previous work such as \citet{kleinberg2018recharging}, the pattern of the supporting points $\{d_i^{(k)}\}$ in Section \ref{ssec:offline-upper-bound} is not controllable in our general setting even for one single arm. $R$ and $R^{cc}$ only coincide on these supporting points, and there can be substantial gaps on non-supporting points. Also, different arms may have very different supporting points, complicating the design of pulling times. To circumvent the challenges, we develop the rounding-scheduling framework in the following. 

\subsubsection*{Step 1: Rounding}


We fix a set $\mathcal D[a] = \bigcup_{j=1}^a \mathcal D_j\cup\{+\infty\}$, where $a$ is a positive integer to be specified later, and where 
    \[
    \mathcal D_j = \{(2j-1)\times 2^{\ell}\}_{\ell\geq 0}.
    \]
We then round each $x_i$ to $1/d_i$ such that $d_i\in\mathcal D[a]$. More precisely, we let $d_i$ be the element in $\mathcal D[a]$ closest to $1/x_i$ while no smaller than $1/x_i$, i.e., 
\begin{align} \label{formula:d_i}
    d_i = \min\{d\geq 1/x_i, d\in\mathcal D[a]\}.
\end{align}
Note that when $x_i=0$, $d_i$ naturally equals to $+\infty$, meaning the arm is never pulled. We have the following lemma that lower bounds the average reward after rounding. 
\begin{lemma} \label{lemma:rounding-lower-bound} If $1/x_i\in\{d_i^{(k)}\}_{k\geq 1}$, then
\begin{align*}
    \frac{R_i(d_i)/d_i}{F_i(x_i)} \geq \frac{a}{a+1}.
\end{align*}
\end{lemma}

\subsubsection*{Step 2: Scheduling}

In this step, we construct a $K$-PPP based on $\{d_i\}$ obtained in Step 1. To achieve this, we first relax the hard constraint $K$ to some positive integer $K[a]\geq K$, which means $K[a]$ arms are allowed to be pulled at the same time. This can be fulfilled with the help of the following lemma.

\begin{lemma} \label{lemma:power-of-2}
    Fix a positive integer $j$. Let $\mathcal I_j = \{i: d_i\in \mathcal D_j\} = \{i_1, i_2, \cdots, i_{|\mathcal I_j|}\}$ and $K_j = \sum_{i\in\mathcal I_j}1/d_i$. Assume $d_{i_1}\leq d_{i_2}\leq \cdots d_{i_{|\mathcal I_j|}}$.
    \begin{itemize}
    \item If $K_j > 1$, then in $\mathcal O(|\mathcal I_j|)$ time, we can find at most $\lceil K_j\rceil$ disjoint sets $\{\mathcal I_{j_s}\}_s$ such that $\bigcup_{s}\mathcal I_{j_s}=\mathcal I_j$ and $\sum_{i\in\mathcal I_{j_s}}1/d_i \leq 1\ (\forall s)$.
    \item If $K_j\leq 1$, then in $\mathcal O\left(|\mathcal I_j|\log d_{i_{|\mathcal I_j|}}\right)$ time, we can specify a $1$-PPP such that we pull each arm $i\in\mathcal I_j$ at frequency $1/d_i$.
    \end{itemize}
\end{lemma}

Now we elaborate on how to use Lemma \ref{lemma:power-of-2} to achieve our goal. Fix any $j\in[a]$, we split $\mathcal I_j$ into several groups by using the first statement in Lemma \ref{lemma:power-of-2}, where in each group the sum of frequencies is within $1$. Then for each group, we apply the second statement in Lemma \ref{lemma:power-of-2} to construct a $1$-PPP. Repeating over all $j\in[a]$ leads to the construction.

We have two observations from the procedure above. First, each arm $i$ is included in exactly one group and is pulled with frequency $1/d_i$. Second, the number of arms pulled at the same time can be bounded by the total number of groups $K[a]$, which is
\begin{align} \label{formula:K[a]}
    K[a] \triangleq \sum_{j\in[a]}\left\lceil\sum_{i\in\mathcal I_j}1/d_i\right\rceil < \sum_{j\in[a]}\sum_{i\in\mathcal I_j}1/d_i + \sum_{j\in[a]}1 \leq \sum_{i}x_i + a.
\end{align}

Now that we have $K[a]$ groups where for each group we have constructed a feasible $1$-PPP, and each arm appears in exactly one group, we choose $K$ of these groups that obtain the largest long-run average reward. These $K$ $1$-PPPs together constitute a $K$-PPP.

We would like to give some brief remarks before proceeding. Our rounding method (Step 1) only requires that the recovery functions are non-decreasing, and our scheduling method (Step 2) does not require any additional assumptions. These may help generalize potential applications of our technique when dealing with more complicated settings. We summarize the procedure above in Algorithm \ref{alg:framework}.

\bigskip
\begin{breakablealgorithm}
   \caption{Rounding and Scheduling: R-S}
   \label{alg:framework}
\begin{algorithmic}[1]
   \State {\bfseries Input:} $x\in[0, 1]^N, K\in\mathbb Z_+, a\in\mathbb Z_+$
   \State {\bfseries Output:} A feasible $K$-PPP
   \State Let $\mathcal D[a] = \bigcup_{j=1}^{a}\left\{(2j-1)\times 2^\ell\right\}_{\ell\geq 0}\cup\{+\infty\}$.
   \For{$i=1$ {\bfseries to} $N$}
   \State $d_i \gets \min\{d\geq 1/x_i, d\in\mathcal D[a]\}$.
   \EndFor
   \For{$j=1$ {\bfseries to} $a$}
   \State Let $\mathcal I_j = \left\{i:d_i\in\mathcal D_j\right\}$ and $s_j=\lceil\sum_{i\in\cI_j}1/d_i\rceil$.
   \State Apply Lemma \ref{lemma:power-of-2} to construct $1$-PPPs: $\cI_{j, 1}, \cdots, \cI_{j, s_j}$.
   \State Let $\cR(\cI_{j, s}) = \sum_{i\in\cI_{j, s}}R_i(d_i)/d_i$ be the long-run average reward of $\cI_{j, s}\ (\forall s)$.
   \EndFor
   \State Select $K$ $1$-PPPs from $\{\cI_{j, s}\}_{j, s}$ that attain the top $K$ largest long-run average reward.
\end{algorithmic}
\end{breakablealgorithm}
\bigskip

\subsection{Main Results}
\label{ssec:offline:results}

Now suppose $x^*$ is an optimal solution of $(\ref{program:upper-bound})$. We can always assume that $x^*$ satisfies the property stated in Lemma \ref{lemma:upper-bound}. Combining the two steps above, we can see that, if  $x_i^*\in\{1/d_i^{(k)}\}_{k\geq 1}\ (\forall i\in[N])$, then
\begin{align*}
K[a] < \sum_{i\in[N]}x_i + a \leq K + a,
\end{align*}
and thus $K[a]\leq K+a-1$. By tuning $a$ appropriately, the long-run approximation ratio of our policy is lower bounded by
\begin{align*}
    \max_a\left\{\frac{\sum_{i\in[N]}R_i(d_i)/d_i}{\sum_{i\in[N]}F_i(x_i^*)}\cdot\frac{K}{K[a]}\right\} \geq \max_{a\in\mathbb Z_+}\frac{a}{a+1}\cdot\frac{K}{K+a-1},
\end{align*}


In the following, we consider dealing with the general case where not all components of the optimal solution are of form $1/d$. If there is an $x_{i_0}^*=0$, then we just ignore arm $i_0$, and this does not hurt the performance. If there is an $x_{i_0}^*>0$ such that $1/x_{i_0}^*\notin\{d_{i_0}^{(k)}\}_{k\geq 1}$, we first round $x_{i_0}^*$ to $\widetilde x_{i_0}^* = \min\left\{y\geq x_{i_0}^*: 1/y\in\{d_{i_0}^{(k)}\}_{k\geq 1}\right\}$ and then feed $\{x_i^*\}_{i\neq i_0}\cup\{\widetilde x_{i_0}^*\}$ to Step 1. Note that we still have $\widetilde x_{i_0}^*\leq 1/d_{i_0}^{(1)}$, and $F_{i_0}(\widetilde x_{i_0}^*)\geq F_{i_0}(x_{i_0}^*)$. Note again that there exists at most one such $i_0$. All the analysis in Step 1 and Step 2 are valid, except that the upper bound of $K[a]$ increases from $K+a-1$ to $K+a$, which means we allow pulling $K+a$ arms at the same time in Step 2. This is because we can only have the bound
\begin{align*}
    \sum_{i\neq i_0}x_i^* + \widetilde x_{i_0}^* + a < K + 1 + a.
\end{align*}
And the ratio becomes
\begin{align*}
    \max_{a\in\mathbb Z_+}\frac{a}{a+1}\cdot\frac{K}{K+a}\triangleq \gamma_K.
\end{align*}
Algorithm \ref{alg:offline} describes the complete paradigm for the offline problem.

\bigskip
\begin{breakablealgorithm}
   \caption{Offline Purely Periodic Planning}
   \label{alg:offline}
\begin{algorithmic}[1]
   \State {\bfseries Input:} $\{R_i(d)\}_{i\in[N], d\geq 1}$
   \State $a^* = \arg\max_a\frac{aK}{(a+1)(K+a)}$.
   \State Initialize the supporting points $\{d_i^{(k)}\}_{i\in[N], k\geq 0}$. 
   \State Initialize $\{F_i(x)\}_{i\in[N], x\in[0, 1]}$ using Lemma \ref{lemma:F-property}.
   \State Solve (\ref{program:upper-bound}) and obtain an optimal solution $x^*$ that satisfies the property of Lemma \ref{lemma:upper-bound}. 
   \For{$i=1$ {\bfseries to} $N$}
   \State $x_i\gets \min\left\{y\geq x_i^*: 1/y\in\{d_i^{(k)}\}_{k\geq 1}\cup\{+\infty\}\right\}$.
   \EndFor
   \State Run the procedure $\text{R-S}(x, K, a^*)$ and obtain a $K$-PPP.
\end{algorithmic}
\end{breakablealgorithm}
\bigskip


Theorem \ref{theorem:offline} provides the theoretical performance of Algorithm \ref{alg:offline}, showing that its long-run approximation ratio is lower bounded by $\gamma_K$.
\begin{theorem} \label{theorem:offline}
For any $T\geq 1$, the total reward of the schedule returned by Algorithm \ref{alg:offline} is lower bounded by
\begin{align*}
    \gamma_K\cdot\UB_{\cR}[K]\cdot T - \mathcal O(N),
\end{align*}
where
\begin{align*}
    \gamma_K = \max_{a\in\mathbb Z_+}\frac{a}{a+1}\cdot\frac{K}{K+a}.
\end{align*}
\end{theorem}

We give some remarks on Theorem \ref{theorem:offline}. First, the term $\cO(N)$ appears because for each arm, we may incur loss at the beginning or the end of the whole time horizon. Second, the long-run approximation ratio $\gamma_K$ satisfies $3$ nice properties: (i) Independence with $N$ and $T$; (ii) Uniformly lower bounded by $1/4$; (iii) Asymptotically approaches $1$ with a gap of $\cO(1/\sqrt{K})$. (iii) holds because the optimal $a$ is either $\lfloor\sqrt{K}\rfloor$ or $\lceil\sqrt{K}\rceil$, and as a result, $1-\gamma_K = \cO(1/\sqrt{K})$.


Then, a natural question arises: Is the gap of $\cO(1/\sqrt{K})$ asymptotically tight with regard to $K$? In our previous discussion, we obtain an offline algorithm that achieves a long-run approximation ratio $\gamma_K$ such that it approaches $1$ with a $\cO(1/\sqrt K)$ gap. Now we give a confirmatory answer to the question \textit{within the class of PPP}.

\begin{theorem} \label{thm:offline-upper-bound}
Fix some $K\geq 2$. Then there exists an instance $\{R_i(d)\}_{i\in[N], d\in\mathbb Z_+}$ such that for any $K$-PPP where arm $i$ is pulled every $d_i$ times, the following holds.
\begin{align*}
    \frac{\sum_{i}R_i(d_i)/d_i}{\UB_{\cR}[K]} \leq 1 - \frac{1}{6\sqrt{K\ln K}+3} + \frac{1}{2K}.
\end{align*}
\end{theorem}

We give some remarks on Theorem \ref{thm:offline-upper-bound}. The bound suggests that the long-run ratio within the PPP class compared to the upper bound is no more than $1-\Omega(1/\sqrt{K\ln K})$ in the worst case. Combined with Theorem \ref{theorem:offline}, we can see that the $\widetilde\Theta(1/\sqrt K)$ gap is asymptotically tight, and our algorithm in Section \ref{ssec:offline-schedule} achieves this tight gap.

Before proceeding, we would like to discuss the relations between our results and the periodic scheduling literature. Our work differs from previously studied periodic scheduling problems in that there is no hard constraint between consecutive occurrences of a job; instead, through our analysis of $\UB[N, K]$ and design of $K$-PPP, it turns out that a carefully designed PPP is able to achieve near-optimal behavior. Our rounding method is related with that in \citet{sgall2009periodic} where jobs are also scheduled in a purely periodic way, but has two main differences. First, the optimal solution to (\ref{program:upper-bound}) may have a component of non-supporting point, which is not faced in typical periodic scheduling problems. This adds some technical difficulties in our setting. Second, since we allow pulling $K$ arms, our technique allows rounding components to numbers in different $\mathcal D_j$. As a result, the ratio $\gamma_K$ tends to 1 as $K$ grows. This is in contrast to selecting one single $\mathcal D_j$ as in \citet{sgall2009periodic}. Our scheduling framework has some relation with that in \citet{holte1989pinwheel}. The class $\mathcal D_j$ in Lemma \ref{lemma:power-of-2} actually belongs to $C_M$ in Section 3 of \citet{holte1989pinwheel}. Nevertheless, there are two different highlights in Lemma \ref{lemma:power-of-2}. First, the first bullet gives a more general picture of efficiently dealing with the situation when the sum of frequencies $> 1$ during the scheduling process, which may happen even when $K=1$. Second, in the proof of the second bullet, we give a more computationally efficient scheduling method for our case, while the value of $m$ in SimpleGreedy of \citet{holte1989pinwheel} may appear quite large. 

\section{The Online Learning Problem} \label{sec:online}

In this section, we turn to address the online counterpart of the offline planning problem. In the online problem, unlike in Section \ref{sec:offline}, the recovery function $\{R_i(d)\}_{i, d}$ is not known a priori and should be learned from the sequential samples for any $i$. 
Our goal is to construct an online learning policy that achieves small regret compared to the offline result in Section \ref{sec:offline}. To be precise, we want to design an online learning algorithm such that the $\gamma_K$-regret, defined by the difference between $\UB_{\cR}[K]$ and the expected total reward obtained by running the algorithm through a time horizon of $T$, is sub-linear in $T$ for general $\cR$ satisfying Assumption \ref{assumption:mono} and \ref{assumption:bound}. Furthermore, we hope that the algorithm is \textit{adaptive}, in the sense that it can obtain better performance guarantee when additional assumptions are made on $\cR$, particularly if the offline planning problem admits a (long-run) approximation ratio better than $\gamma_K$. We discuss our results under only Assumption \ref{assumption:mono} and \ref{assumption:bound}.

\subsection{Design Framework}

Broadly speaking, our policy design is built upon the ``optimism under uncertainty'' principle that has been successfully applied in many online learning problems. However, there are several main difficulties we have to confront in the recovering setting.

\begin{enumerate}
\item $\{R_i(d)\}$ is in essense non-parametric, and a complete sample of $R_i(d)$ requires at least $d+1$ time periods since we have to wait for the delay and plan for $d$ time periods in the future. This planning issue is not faced in classical stochastic bandit problems.
\item Due to sampling error, the upper confidence bounds of the estimation of $\{R_i(d)\}_{d\geq 1}$ may not be non-decreasing, i.e., violating Assumption \ref{assumption:mono}. This impedes us from directly plugging the upper confidence bounds into $F_i(\cdot)$.
\item To make things more complicated, we have no prior knowledge of $\{d_i^{(k)}\}_{i\in[N], k\geq 1}$, which is crucial for estimating $\{F_i(\cdot)\}$. In previous work such as \citet{kleinberg2018recharging} and \citet{basu2019blocking}, $\{d_i^{(k)}\}_{i\in[N]}$ are always known a priori. Under Assumption \ref{assumption:mono} and \ref{assumption:bound}, $\{R_i(d)\}_{d\geq 1}$ may appear to be an irregular shape, causing trouble for recovering $F_i(\cdot)$.
\end{enumerate}

To address the first difficulty, we divide the whole time horizon into several phases of length $\phi$. At the beginning of each phase $j$, we run an offline oracle to plan for a schedule in that phase. At the end of each phase, we update our estimation of $\{R_i(d)\}$, construct its corresponding set of UCBs $\{\hat R_{i, j}(d)\}$, and feed it to the offline oracle for planning the next phase $j+1$. We note that $\phi$ has to be carefully tuned because there is a trade-off: If $\phi$ is too large, we might explore too much on a sub-optimal schedule in some phase. While if $\phi$ is too small, we cannot estimate $R_i(d)$ for some larger $d$, causing us to get stuck in a sub-optimal schedule.

The UCBs are constructed as follows. Given a phase $j$, for each $i\in[N]$ and $d\geq 1$, we let $n_{i, j-1}(d)$ be the number of samples we have collected for $R_i(d)$ prior to phase $j$, and $\bar R_{i, j-1}(d)$ be the empirical mean of $R_i(d)$ prior to phase $j$. We let $n_{i, 0}(d)=0$ and $\bar R_{i, 0}(d) = 0$. Then the upper confidence bound of $R_i(d)$ is
   \begin{align} \label{formula:UCB}
   \hat R_{i, j}(d) & \triangleq \min\left\{\bar R_{i, j-1}(d) + R_{\max}\sqrt{\frac{2\log (KT)}{\max\left\{n_{i, j-1}(d), 1\right\}}}, \ R_{\max}\right\}.
   \end{align}

To address the second and third difficulty, we need to re-interpret what Section \ref{sec:offline} implies us on constructing a proper offline oracle. A crucial observation is as follows: The approximation ratio $\gamma_K$ for the offline problem is achieved (even) if we (i) confine ourselves only to the class of PPP, and meanwhile (ii) restrict the possible periods to some (sparse) set $\mathcal D[a]$ with appropriate $a$. This observation leads to the offline oracle we describe as follows. 

Fix a phase $j$ and a set of integers $\cD\subset\mathbb Z_+$. Let $\{\hat R_{i, j}(d)\}_{i\in[N], d\in\cD_\phi[a]}$ be some (estimated) UCBs over $\{R_i(d)\}_{i\in[N], d\in\cD_\phi[a]}$. Let $K'\geq 0$ be a non-negative number. Let $x_{i, j, d}$ denote whether we pull arm $i$ with period $d\in\cD_\phi[a]$ in phase $j$. Consider the following problem:
\begin{align} \label{program:knapsack}
    \max_x \quad & \sum_{i\in[N]}\sum_{d\in\cD}\hat R_{i, j}(d)x_{i, j, d}/d \\
    \text{s.t.} \quad & \sum_{i\in[N]}\sum_{d\in\cD}x_{i, j, d}/d\leq K', \nonumber\\
    & \sum_{d\in\cD}x_{i, j, d}\leq 1, \quad \forall i\in[N], \nonumber\\
    & x_{i, j, d}\in\{0, 1\}, \quad \forall i\in[N], d\in\cD. \nonumber
\end{align}
In the following we discuss how to choose $\cD$. Let $a\in\mathbb Z_+$. Define $\mathcal D_\phi[a] = \{d: d\leq \phi/2, d\in\mathcal D[a]\}$ as the set of periods that are included in $\mathcal D[a]$ but at the meantime no larger than $\phi/2$. This is imposed such that in each phase all selected periods can be estimated at least once. We give some explanations on (\ref{program:knapsack}) with $\cD = \cD_\phi[a]$. It can be regarded as a generalization of the knapsack problem that combines solving (\ref{program:upper-bound}) in Section \ref{ssec:offline-upper-bound} with Step 1 in Section \ref{ssec:offline-schedule}. We seek to maximize the long-run average reward with a frequency constraint, where $K'$ is a frequency parameter. In the knapsack problem, each item has two choices: to be selected or not. While in (\ref{program:knapsack}), an arm has different versions indexed by $d\in\cD$, and we can choose at most one of the versions for each arm. Note that to implement (\ref{program:knapsack}), it's sufficient to collect samples for periods only in $\cD$ rather than the overall positive integers.

We now address the computation issue. (\ref{program:knapsack}) is an integer programming, and is in general not polynomial-solvable. Nevertheless, similar to the knapsack problem, (\ref{program:knapsack}) admits an efficient FPTAS shown in Lemma \ref{lemma:knapsack}. 

\begin{lemma} \label{lemma:knapsack}
Let $\cD=\cD_\phi[a]$. We can obtain a $(1-\epsilon)$-optimal solution of (\ref{program:knapsack}) in $\mathcal O\left(\frac{N^3a\log_2\phi}{\epsilon}\right)$ time.
\end{lemma}

Once we obtain a $(1-\epsilon)$-optimal solution $\{x_{i, j, d}^\epsilon\}$ of (\ref{program:knapsack}), we let
\begin{align*}
    x_{i, j}^\epsilon = \sum_{d\in\mathcal D_\phi[a]}x_{i, j, d}^\epsilon/d
\end{align*}
be the frequency of pulling arm $i$. The next stage is to construct a $K$-PPP. Note that $\{x_{i, j}^\epsilon\}_{i\in[N]}$ itself does not necessarily constitute the set of frequencies of a feasible $K$-PPP because it only satisfies the frequency constraint. We feed $\{x_{i, j}^\epsilon\}_{i\in[N]}$ into the R-S procedure described in Section \ref{ssec:offline-schedule} to obtain feasible periods $\{d_{i, j}\}$ and starting times $\{t_{i, j}\}$ which together constitute a $K$-PPP in phase $j$. Finally, we run this $K$-PPP till the end of phase $j$ and proceed to phase $j+1$ if we are not at the final time period.

Algorithm \ref{alg:online} describes the complete procedure for the online learning problem. We would like to note that $a, K'$ are two tuning parameters designed for flexibility. They can be chosen adaptively according to different scenarios. We will illustrate this point in the next section.

\bigskip
\begin{breakablealgorithm}
   \caption{Online Purely Periodic Learning}
   \label{alg:online}
\begin{algorithmic}[1]
   \State {\bfseries Input:} phase length $\phi$, computational accuracy $\epsilon$, tuning parameters $a$, $K'$.
   \State $t\gets 0$, $j\gets 1$.
   \State $\cD\gets \mathcal D_\phi[a] = \{d: d\leq \phi/2, d\in\mathcal D[a]\}$.
   \Repeat
   \State Construct UCBs $\{\hat R_{i, j}(d)\}_{i\in[N], d\in\cD_\phi[a]}$ as in (\ref{formula:UCB}).
   \State Solve (\ref{program:knapsack}) with UCBs and obtain a $(1-\epsilon)$-optimal solution $\{x_{i, j, d}^\epsilon\}_{i\in[N], d\in\cD_\phi[a]}$ by Lemma \ref{lemma:knapsack}.
   \State Let $x_i = \sum_{d\in\cD_\phi[a]}x_{i, j, d}^\epsilon/d\ (\forall i\in[N])$.
   \State Run $\text{R-S}(x, K, a)$ and obtain a $K$-PPP with the set of periods $\{d_{i, j}\}$ and the set of starting times $\{t_{i, j}\}$.
   \State Run this $K$-PPP for $\min\{\phi, T-t\}$ time periods.
   \State $t\gets t + \phi$, $j\gets j + 1$.
   \Until{$t \geq T$}
\end{algorithmic}
\end{breakablealgorithm}
\bigskip

\subsection{Theoretical Guarantee}

In this section, we carry out analysis on Algorithm \ref{alg:online}. Denote $\val_\cR[K'|\cD]$ as the optimal objective value of (\ref{program:knapsack}) when $\hat R_{i, j}(d) = R_i(d)$ for all $i\in[N]$ and $d\in\cD_\phi[a]$ (that is, the coefficients in (\ref{program:knapsack}) are replaced by the true rewards). Theorem \ref{theorem:online} states a lower bound of the total reward obtained from Algorithm \ref{alg:online}.

\begin{theorem} \label{theorem:online}
Let $\cD = \cD_\phi[a]$. Fix $\phi, \epsilon, a, K'$ as the input parameters for Algorithm \ref{alg:online}, then the expected overall reward achieved by Algorithm \ref{alg:online} is at least
\begin{align}
    & \min\left\{1, \frac{K}{K'+a-1}\right\}(1-\epsilon)\cdot \val_\cR[K'|\cD]\cdot T - \nonumber\\
    & \quad C\cdot R_{\max}\left(\phi N\log (a+1)\sqrt{\log (KT)} + \sqrt{aKNT\log\phi\log (KT)} + \frac{NT}{\phi}\right), \label{formula:regret}
\end{align}
where $C>0$ is an absolute constant.
\end{theorem}

We give some remarks on Theorem \ref{theorem:online}. The first line of (\ref{formula:regret}) is the \textit{benchmark} term. The factor $\frac{K}{K'+a-1}$ is incurred because of the scheduling step in the R-S procedure. $\epsilon$ is only a parameter controlling the accuracy of solving (\ref{program:knapsack}). $\val_\cR[K'|\cD]$ is an upper bound of the long-run average reward if we confine ourselves to PPP with periods in $\cD_\phi[a]$. One advantage of using $\val_\cR[K'|\cD]$ is its flexibility, in the sense that any theoretical improvement on the offline ratio guarantee of our framework can be directly transformed into an online result. We will illustrate this point and show the connection to $\UB_\cR[K]$ in our subsequent discussions for both general cases and special ones. 

The second line of (\ref{formula:regret}) is the \textit{regret} term consisting of three components. The first component is the \textit{initialization loss}. The intuition is that each arm has to be explored at least once, and the loss incurred by exploring one arm is approximately $\tilde\cO(\phi)$ because there are $\phi$ time periods in each phase. The second term is the loss incurred by standard exploration-exploitation. We note that the design of (\ref{program:knapsack}) serves as a crucial role in the proof of this second term. Compared to traditional MAB problems, we only pay an additional factor of $a\log \phi$ since we only need to explore periods within a sparse set on $\mathbb Z_+$ for each arm by using (\ref{program:knapsack}). Such construction is implied by our offline result in Section \ref{sec:offline}, which in turn relies on the monotonicity assumption (Assumption \ref{assumption:mono}). The third component is the \textit{transition loss}, which is incurred by the transition of planning schedules between each two adjacent phases. Since the length of each phase is $\phi$, there are $\cO(\frac{T}{\phi})$ transitions. For each transition, the loss is $\cO(NR_{\max})$ compared to the long-run upper bound $\val_\cR[K'|\cD]$.

We leave the detailed logarithm terms in the regret bound to the supplementary material. Note that in Theorem \ref{theorem:online}, $\epsilon\in(0, 1]$ is only a parameter for computational accuracy of solving (\ref{program:knapsack}). It is not related with the regret bound.


In the following, we analyze the properties of (\ref{program:knapsack}). Thanks to Assumption \ref{assumption:bound}, we have Lemma \ref{lemma:online-offline-lower} that relates the objective value of (\ref{program:knapsack}) to that of (\ref{program:upper-bound}).

\begin{lemma} \label{lemma:online-offline-lower}
Let $K'=K+1$ and $\cD=\cD_\phi[a]$. Then we have
\begin{align*}
    \val_{\cR}[K'|\cD]\geq\frac{a}{a+1}\UB_{\cR}[K] - \frac{2NR_{\max}}{\phi}.
\end{align*}
\end{lemma}

The remaining issue is to tune $a$ and $\phi$ appropriately. Combined with Theorem \ref{theorem:online}, we have the following corollary.
\begin{corollary} \label{coro:online-general}
Let the parameters be defined as follows, 
\begin{align*}
    & \phi=\Theta\left(\sqrt{\frac{T}{\log (K+1)}}\right), \quad \epsilon=\Theta(1/\sqrt{T}), \\
    & a = \arg\max_{a'}\frac{a'K}{(a'+1)(K+a')}, \quad K' = K+1.
\end{align*}
Then the expected overall reward achieved by Algorithm \ref{alg:online} is at least
\begin{align*}
    \gamma_K\UB_{\cR}[K]\cdot T - \widetilde\cO \left(\max\{N, K^{\frac{3}{4}}N^{\frac{1}{2}}\}\sqrt T\right).
\end{align*}
\end{corollary}

Some remarks need to be addressed. First, the result stated in Corollary \ref{coro:online-general} gives a performance guarantee for general reward recovering instances. Note that the \textit{benchmark} term $\gamma_K\UB_\cR[K]T$ is the same as that in Theorem \ref{theorem:offline}. $\widetilde\cO \left(\max\{N, K^{\frac{3}{4}}N^{\frac{1}{2}}\}\sqrt T\right)$ is the \textit{regret} term. It is the best we can hope since in many MAB problems, a $\widetilde \cO(\sqrt T)$ regret is minimax optimal with respect to $T$. As to the computation issue, since $\epsilon=\Theta(1/\sqrt{T})$, from Lemma \ref{lemma:knapsack}, the overall computation throughout the whole time horizon is bounded by $\widetilde\cO(\sqrt K N^3T)$. In practice, the four parameters in Algorithm \ref{alg:online} can be tuned to achieve better empirical performance.

The second remark is the \textit{adaptiveness} of Algorithm \ref{alg:online}. We can possibly improve the theoretical performance for special classes of $\cR$. The reason is that $\val_\cR[K'|\cD]$ may have a better bound than that in Lemma \ref{lemma:online-offline-lower} if additional assumptions are made on $\cR$. We will demonstrate this point in more details in Section \ref{sec:worst-case}. An example is the standard combinatorial MAB problem without the recovering behavior where in each time period at most $K$ arms can be pulled. If we let $K'=K$ and $a=1$, then we have an improved version of Lemma \ref{lemma:online-offline-lower}:
\begin{align*}
    \val_\cR[K'|\cD]\geq\UB_{\cR}[K] - \frac{2NR_{\max}}{\phi}.
\end{align*}
With $\phi=\Theta(\sqrt T)$, by Theorem \ref{theorem:online}, the expected overall reward achieved by Algorithm \ref{alg:online} is then at least
\begin{align*}
    \UB_\cR[K]\cdot T - \widetilde\cO (N\sqrt T).
\end{align*}
Compared to existing literature, we have an additional $\sqrt{N/K}$ factor in the regret term because of the \textit{transition loss} in our bound.


\section{Refining the Worst-case Ratio} \label{sec:worst-case}

In Section \ref{sec:PPP} and Section \ref{sec:online}, we have obtained worst-case ratios that stand for arbitrary $K$ through a unified framework. The worst-case ratio tends to one with a $O(1/\sqrt{K})$ gap as $K$ increases to infinity. When $K=1$, the ratio stands as $1/4$. In this section, we establish a new routine under the unified framework, in terms of both algorithm design and theoretical analysis, to enhance this worst-case ratio for small $K$. One implication of the results in this section is that the offline and online algorithms proposed in Section \ref{sec:offline} and \ref{sec:online}, which work for arbitrary $K$, can be further improved if some real-world problems give rise to small $K$. We emphasize again that the discussion still lies in the problem framework in Section \ref{sec:offline}. We do not make any additional assumptions on the reward functions other than monotonicity and boundedness.


The road-map of this section is as follows. We first extend Algorithm \ref{alg:offline} to allow for more flexible methods of rounding as well as dealing with the non-supporting component to schedule a PPP. These combined methods are integrated into one algorithm and can be proven to obtain a $1/2$ worst-case ratio. Meanwhile, we show that $1/2$ is not improvable for general reward instances within our design paradigm. Then we address the online learning problem by utilizing the offline results and constructing a computationally efficient offline oracle.

We now re-state some preliminaries that will be useful throughout the discussions in this section. Let $\cD_{a} = \{(2a-1)\times 2^\ell\}_{\ell\geq 0}$ be a subset of integers indexed by $a$. By Lemma \ref{lemma:upper-bound}, we can assume that the optimal solution $x^*$ to (\ref{program:upper-bound}) satisfies the property stated in Lemma \ref{lemma:upper-bound}. That is, there is at most one component of $x^*$ that appears to be a non-unit fraction. Without loss of generality, we assume that the component is $x_1^*$. Then there exists some $\alpha\in[0, 1]$ and $k_1\geq 1$ such that
\begin{align*}
    x_1^* & = \alpha\frac{1}{d_1^{(k_1)}} + (1-\alpha)\frac{1}{d_1^{(k_1+1)}} > 0, \\
    F_1(x_1^*) & = \alpha\frac{R_1(d_1^{(k_1)})}{d_1^{(k_1)}} + (1-\alpha)\frac{R_1(d_1^{(k_1+1)})}{d_1^{(k_1+1)}}.
\end{align*}
Then for any $i>1$, we have $1/x_i^*\in\{d_i^{(k)}\}_{k\geq 1}\cup\{+\infty\}$. 

\subsection{Offline Planning}

Before introducing the formal algorithm, we present some intuitions on why we can improve the ratio from $1/4$ to a larger (better) constant. 

\begin{itemize}
    \item The analysis for Algorithm \ref{alg:offline} may not be tight. In our analysis in Section \ref{sec:offline}, both the rounding error as well as the scheduling error can be controlled by $1/2$, and thus the worst-case ratio is guaranteed by
    \begin{align*}
        \left(1-\frac{1}{2}\right)\cdot\left(1-\frac{1}{2}\right) = \frac{1}{4}.
    \end{align*}
    However, the analysis still has room to improve because we considered the two types of error separately in Section \ref{sec:offline}. In fact, when one error is large, the other one may turn out to be small. \textit{Jointly} bounding the two types of error may help improve the worst-case ratio.
    \item There may exist other methods to deal with the non-supporting component. In Algorithm \ref{alg:offline}, we first do
    \begin{align} \label{formula:non-unit-1}
        x_i = \min\left\{y\geq x_i^*: 1/y\in\{d_i^{(k)}\}_{k\geq 1}\right\} \geq x_i^*\quad \forall i\in[N].
    \end{align}
    Note that $x_i = x_i^*\ (\forall i > 1)$, $x_1 = 1/d_1^{(k_1)}$ and $1/x_i\in\{d_i^{(k)}\}_{k\geq 1}\ (\forall i\in[N])$. The resulting (long-run) average reward is
    \begin{align*}
        \sum_{i\in[N]}R_i(1/x_i)x_i = \sum_{i\in[N]}F_i(x_i) \geq \sum_{i\in[N]}F_i(x_i^*).
    \end{align*}
    This means the upper bound we are comparing with is actually larger than $\UB_\cR[1]$ because $R_1(d_1^{(k)})/d_1^{(k)} \geq F_1(x_1^*)$. Such inflation in the objective value may benefit the worst-case ratio. On the other hand, if this inflation is small, then even if enlarging $x_1^*$ as in Algorithm \ref{alg:offline} may not be beneficial, shrinking $x_1^*$ may not lose much. Notice that one advantage of shrinking $x_1^*$ is that the scheduling error may become smaller.
    \item Periods may not necessarily be set as power of $2$. In Algorithm \ref{alg:offline}, when $K$ is small ($K\leq 6$ to be precise), we will always choose $a=1$. That is, we only adopt periods that are power of 2. However, we can also choose $a > 1$ and run the R-S procedure with $\cD[a]$ replaced by $\cD_a$. This will possibly improve the worst-case ratio. For example, when $1/x_i=2^\ell+1\ (\ell > 0)$, $d_i=2^{\ell+1}$ if we choose $a=1$, while $d_i=3\times 2^{\ell-1}$ if we choose $a=2$. Apparently, letting $a=2$ will be more beneficial under the circumstance because we lose less when doing rounding.
\end{itemize}



We now present Algorithm \ref{alg:offline-1} and elaborate on its components.

\bigskip
\begin{breakablealgorithm}
   \caption{Offline Purely Periodic Planning}
   \label{alg:offline-1}
\begin{algorithmic}[1]
   \State {\bfseries Input:} $x^*\in[0, 1]^N$ that satisfies the property of Lemma \ref{lemma:upper-bound}
   \State {\bfseries Output:} A feasible $K$-PPP
   \For{$a=1, 2, 3$}
   \For{$m = 1, 2, 3$}
   \For{$i=1$ {\bfseries to} $N$}
   \State
   \[
   \left\{
   \begin{array}{lc}
   x_i \gets \min\left\{y\geq x_i^*: 1/y\in\{d_i^{(k)}\}_{k\geq 1}\cup\{+\infty\}\right\}, & \quad\text{if }m = 1 \\
   x_i \gets x_i^*, & \quad\text{if }m = 2 \\
   x_i \gets \max\left\{y\leq x_i^*: 1/y\in\{d_i^{(k)}\}_{k\geq 1}\cup\{+\infty\}\right\}, & \quad\text{if }m = 3
   \end{array}
   \right.
   \]
   \State $d_i \gets \min\left\{d\geq 1/x_i, d\in\{(2a-1)\times 2^\ell\}_{\ell\geq 0}\cup\{1, +\infty\}\right\}$.
   \EndFor
   \State Run $\text{R-S}(x, K, a)$ and obtain a $K$-PPP $\cI_{a, m}^*$.
   \EndFor
   \EndFor
\State Select $\cI^* = \arg\max_{a, m}\{\cR(\cI_{a, m}^*)\}$.
\end{algorithmic}
\end{breakablealgorithm}
\bigskip

We first fix some $a\in\{1, 2, 3\}$, and then all positive periods will only appear as the form of $(2a-1)\times 2^\ell$ or $1$. Then we try to fix the issue caused by the non-supporting component, which is the main obstacle for improving the ratio. Apart from the method stated in (\ref{formula:non-unit-1}), we propose two other methods of dealing with the non-supporting component. More precisely, the first method ($m=1$) is the same as that adopted in Algorithm \ref{alg:offline} as well as in (\ref{formula:non-unit-1}). That is, we first enlarge $x_1^*$ to $1/d_1^{(k_1)}$ and then do the R-S procedure. In the second method ($m=2$), we do not make any changes to $x_1^*$ prior to the rounding step. That is, we directly feed $x^*$ into the R-S procedure. In the third method ($m=3$), we first shrink $x_1^*$ to $1/d_1^{(k_1+1)}$ and then do the R-S procedure. 
For each $a$ and each $m$, we can obtain a feasible $K$-PPP. Finally, the algorithm choose the best performed policy among the ($3\times 3=$)$9$ $K$-PPPs.

Lemma \ref{lemma:bound-of-method} gives lower bounds for the (long-run) average reward of the best $K$-PPP for each fixed $m\in\{1, 2, 3\}$. The ``best'' is obtained over different choices of the set of periods $\cD_a$. When $m=1$, we relate the lower bound to the \textit{local} average reward of arm $1$. When $m=2$ or $m=3$, the lower bound is concerned with the \textit{global} average reward of all arms. 

\begin{lemma} \label{lemma:bound-of-method}
We always have
\begin{gather*}
    \max_{a} \cR(\cI_{a, 1}^*) \geq \frac{3}{4}R_1(d_1^{(k_1)})/d_1^{(k_1)}, \\
    \max_{a} \cR(\cI_{a, 3}^*) \geq \frac{3}{5}\left(R_1(d_1^{(k_1+1)})/d_1^{(k_1+1)} + \sum_{i>1}F_i(x_i^*)\right).
\end{gather*}
If $x_1^*\leq 1/2$, we also have
\begin{align*}
    \max_{a} \cR(\cI_{a, 2}^*) \geq \frac{3}{5}\left(R_1(d_1^{(k_1)})x_1^* + \sum_{i>1}F_i(x_i^*)\right).
\end{align*}
\end{lemma}

Now we are ready to present the results for the worst-case ratio guarantee. Theorem \ref{prop:offline} states that the $K$-PPP returned from Algorithm \ref{alg:offline-1} yields a (long-run) $1/2$ worst-case ratio.
\begin{proposition} \label{prop:offline}
For any $T\geq 1$, the total reward of the schedule returned by Algorithm \ref{alg:offline-1} is lower bounded by
\begin{align*}
    \frac{1}{2}\cdot\UB_{\cR}[K]\cdot T - \mathcal O(N),
\end{align*}
\end{proposition}

Before ending the offline planning part, we give a remark on the tightness of the ratio $1/2$. We claim that $1/2$ is the best we can hope if we only consider $K$-PPP's in which there exists some $a\geq 1$ such that the period of any arm $i$ is in $\cD_a\cup\{1, +\infty\}$. Apparently, the $K$-PPP returned by Algorithm \ref{alg:offline-1} falls into the category described described above.

Let $K=1$, $N=2$ and $\ell \geq 0$. Consider the instance $\cR$ as follows. The first arm satisfies 
\begin{align*}
    R_1(d) = 1, \quad \forall d\geq 1.
\end{align*}
The second arm satisfies
\begin{align*}
    R_2(d) = (2^\ell+1)\mathds 1\left\{d\geq 2^\ell+1\right\}, \quad \forall d\geq 1.
\end{align*}
The optimal schedule is to pull arm $2$ once every $2^\ell + 1$ times and pull arm $1$ in the remaining times. The (long-run) average reward equals
\begin{align*}
    \frac{1}{2^\ell+1}(2^\ell+1 + 2^\ell) = 2-\frac{1}{2^\ell + 1}.
\end{align*}
Let's upper bound the average reward of any $K$-PPP for this instance. If there exists an arm $i\in\{1, 2\}$ such that $d_i=1$. Then since $K=1$, we will always pull arm $i$. The (long-run) average reward is no larger than $1$ (we always pull arm $1$). Now we fix any $a\geq 1$ and assume that $d_i > 1\ (\forall i\in\{1, 2\})$. We differentiate between two cases. 

Case 1: $a=1$. The (long-run) average reward can be bounded as
\begin{align*}
    \frac{1}{d_1} + \frac{(2^\ell+1)\mathds 1\left\{d\geq 2^\ell+1\right\}}{d_2} \leq \frac{1}{2} + \frac{2^\ell+1}{2^{\ell+1}} = 1 + \frac{1}{2^{\ell+1}}.
\end{align*}

Case 2: $a>1$. Then if $d_2 > 2^\ell$, we must have
\begin{align*}
    \frac{2^\ell}{d_2} \leq \frac{2a-2}{2a-1},
\end{align*}
since $d_2$ has an odd factor of $2a-1$. Therefore, the (long-run) average reward can be bounded as
\begin{align*}
    \frac{1}{d_1} + \frac{(2^\ell+1)\mathds 1\left\{d\geq 2^\ell+1\right\}}{d_2} \leq \frac{1}{2a-1} + \frac{2^\ell\mathds 1\left\{d_2\geq 2^\ell+1\right\}}{d_2} + \frac{1}{2^\ell + 1} \leq 1 + \frac{1}{2^\ell+1}.
\end{align*}
Wrapping up the analysis above, we can see that for arbitrary $\ell > 0$, the ratio of $K$-PPP can be no larger than
\begin{align*}
    \frac{1+\frac{1}{2^\ell+1}}{2-\frac{1}{2^\ell+1}}.
\end{align*}
Taking $\ell\to+\infty$, the worst-case ratio is then upper bounded by $1/2$.


\subsection{Online Learning}

After obtaining the $1/2$ ratio in the offline problem, we present the online learning result. 

\bigskip
\begin{breakablealgorithm}
   \caption{Online Purely Periodic Learning}
   \label{alg:online-1}
\begin{algorithmic}[1]
   \State {\bfseries Input:} phase length $\phi$.
   \State $t\gets 0$, $j\gets 1$.
   \For{$1\leq a\leq 3$}
   \State $\cD_{a, \phi} \gets \{d: d\in \cD_a, d\leq \frac{\phi}{2}\}$.
   \EndFor
   \Repeat
   \State Construct UCBs $\{\hat R_{i, j}(d)\}_{i\in[N], d\in\bigcup_{1\leq a\leq 3}\cD_{a, \phi}}$ as in (\ref{formula:UCB}).
   \State Let $a^* = \arg\max_{1\leq a\leq 3}\ \val[K|\cD_{a, \phi}]$ and $\{x_{i, j, d}\}_{i\in[N], d\in\cD_{a^*, \phi}}$ be an optimal solution for (\ref{program:knapsack}) with $\cD = \cD_{a^*, \phi}$. Let $x_i = \sum_{d\in\cD_{a^*, \phi}}x_{i, j, d}/d\ (\forall i\in[N])$.
   \State Run $\text{R-S}(x, K, a^*)$ and obtain a $K$-PPP with the set of periods $\{d_{i, j}\}$ and the set of starting times $\{t_{i, j}\}$.
   \State Run this $K$-PPP for $\min\{\phi, T-t\}$ time periods.
   \State $t\gets t + \phi$, $j\gets j + 1$.
   \Until{$t \geq T$}
\end{algorithmic}
\end{breakablealgorithm}
\bigskip

We still apply the online learning framework (Algorithm \ref{alg:online}), but with two main differences. First, rather than obtain a $(1-\epsilon)$-optimal solution to (\ref{program:knapsack}), we can actually solve it to optimal. Again, we take advantage of power of 2 to speed up the computation. We give Lemma \ref{lemma:knapsack-1}, an analogues of Lemma \ref{lemma:knapsack}, stating that (\ref{program:knapsack}) can be solved to optimal in polynomial time.

\begin{lemma'}{lemma:knapsack} \label{lemma:knapsack-1}
Fix $a\geq 1$ and let $\cD=\cD_{a, \phi}$. We can solve (\ref{program:knapsack}) to optimal in $\cO(N\phi\log_2\phi)$ time.
\end{lemma'}

Second, we choose $K'$ and $\cD$ for (\ref{program:knapsack}) different from those in Algorithm \ref{alg:online}. Instead of choosing $\cD = \cD_\phi[a]$, we choose a single series of periods $\cD_{a, \phi}$ each time when solving (\ref{program:knapsack}). Furthermore, we do not ``widen'' the frequency constraint from $K$ to $K+1$ as in Algorithm \ref{alg:online}. The intuitive reason is that regardless of the detailed components in the output $K$-PPP from Algorithm \ref{alg:offline-1}, there always exists some $1\leq a\leq 3$ such that the $K$-PPP pull any arm with period in $\cD_a$. Formally, we have Lemma \ref{lemma:online-offline-lower-1}, an analogues of Lemma \ref{lemma:online-offline-lower} that lower bounds $\val_{\hat\cR}[K'|\cD_{a, \phi}]$ for empirical $\cR$.

\begin{lemma'}{lemma:online-offline-lower} \label{lemma:online-offline-lower-1}
Let $K'=K$. Then we have
\begin{align*}
    \max_{a\in\{1, 2, 3\}}\val_{\cR}[K'|\cD_{a, \phi}] \geq \frac{1}{2}\UB_{\cR}[K] - \frac{2NR_{\max}}{\phi}.
\end{align*}
\end{lemma'}

Now we arrive at Proposition \ref{prop:online} indicating that the regret compared to the offline benchmark is $\widetilde\cO(\sqrt T)$.

\begin{proposition} \label{prop:online}
Let $\phi=\Theta\left(\sqrt{\frac{T}{\log (K+1)}}\right)$. Then the expected overall reward achieved by Algorithm \ref{alg:online-1} can be lower bounded by
\begin{align*}
    \frac{1}{2}\UB_{\cR}[K]\cdot T - \widetilde\cO \left(N\sqrt T\right).
\end{align*}
\end{proposition}

\section{Synthetic Experiments}


In this section, we use synthetic experiments to demonstrate the performance of our algorithms (proposed and analyzed in Section \ref{sec:PPP} to \ref{sec:worst-case}). We note that for the planning problem with a very general recovering reward model and the allowance to pull multiple arms at the same time, existing policies (see, e.g., \citealt{kleinberg2018recharging}, \citealt{papadigenopoulos2021recurrent}) do not have direct generalizations to such planning problem setting except for the greedy policy (see, e.g., \citealt{basu2019blocking}, \citealt{atsidakou2021combinatorial}). Further, existing policies do not have direct generalizations that address the online learning version of our general setting, even for the greedy policy. Therefore, our plan in this section is to demonstrate our proved theoretical guarantees and gain insights on how the performance of our algorithms depend on different parameters. In the offline planning problem, we show through experiments the dependence on the problem parameter $K$. In the online learning problem, we show through experiments the dependence on the tuning parameter $\phi$. 

\subsection{Experiments on Offline Planning} \label{sec:simulation:offline}

\subsubsection*{Settings} We consider four difference choices of $N\in\{50, 100, 150, 250\}$. For any given $N$, we traverse $K$ from $1$ to $N$. For each $(N, K)$, an instance $\cR$ is simulated according to the rule described as follows. For each $i\in[N]$, we sample a $d_i^{\max}$ that is uniformly distributed on $\{1, 2,\cdots, 25\}$ as the maximum recovery time. That is, when $d > d_{i}^{\max}$, $R_i(d)=R_i(d_i^{\max})$. We then sample $d_i^{\max}$ copies of uniform random variables on $[0, 1]$ and sort them by ascending order $u_{i, 1}, \cdots, u_{i, d_i^{\max}}$. We also sample a number $a$ as the absolute value of a logistic distribution. Then we let
\begin{align*}
    R_i(d) = (1+a) \cdot u_{i, d}.
\end{align*}

\subsubsection*{Algorithms} We consider two different classes of policies. The first one is the greedy policy, where in each time period, we always choose the $K$ arms that attain the maximum rewards out of $N$ arms. The second one is the $K$-PPP algorithm proposed in Section \ref{sec:PPP}. To improve the empirical performance especially when $K$ is small, we make two amendments to Algorithm \ref{alg:offline}. One thing is that we separately run $a^*=\lfloor\sqrt K\rfloor$ and $a^*=\lceil\sqrt K\rceil$ instead of letting $a^*=\arg\max_a\frac{aK}{(a+1)(K+1)}$. The other amendment is that we try three different methods of dealing with the non-supporting point (as in Algorithm \ref{alg:offline-1}) instead of only one method (as in Algorithm \ref{alg:offline}, Line 7). With the two amendments above, we will output ($2\times 3=$) $6$ different $K$-PPP's. We also run Algorithm \ref{alg:online-1} that yields $3$ separate results. Finally, we select the $K$-PPP that attains the largest long-run average reward among the $(6+3=)$ 9 results. 

\subsubsection*{Implementation Details}
For each $(N, K)$ pair, we sample $50$ instances through the procedure described above and run the algorithms to compute the long-run average reward. For each instance, we compute the ratio between the average reward and the upper bound $\UB_\cR[K]$. We then compute the ratio between the two metrics and average them over the $50$ instances. However, we note that it is difficult to exactly compute the long-run average reward of the greedy policy. Therefore, in each of our test, we run the greedy policy for $T=10000$ time periods and calculate the (long-run) average reward over the $T$ time periods.

\subsubsection*{Results}

\begin{figure}[!ht]
    \centering
    \includegraphics[width=16cm]{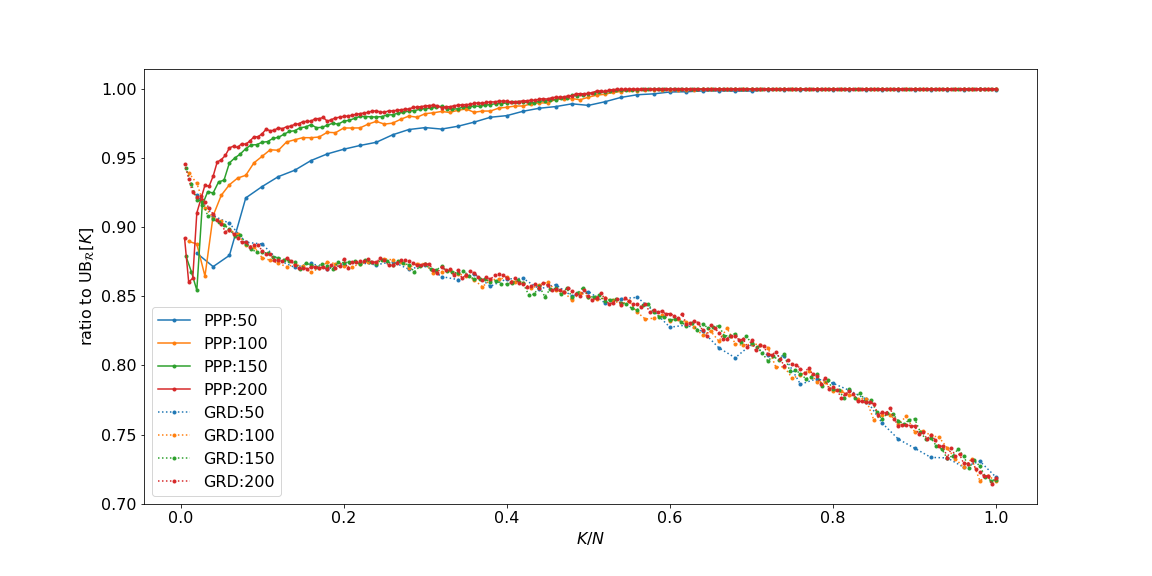}
    \caption{Ratio compared to true optimal for different algorithms}
    \label{fig:offline}
\end{figure}

We plot the ratios in Figure \ref{fig:offline}. There are $8$ lines in total. Each line corresponds to the ratios for fixed policy (``GRD'' means the greedy policy) and $N$ with various $K\in[1, N]$. To make the results more comparable, we make a normalization and take $K/N\in(0, 1]$ as the $x$ axis. Some observations are as follows:
\begin{enumerate}
    \item The greedy policy performs relatively well when $K$ is small, but deteriorates as $K$ grows. Somewhat interestingly, the decaying shape exhibits similar among different $N$'s. The performance decays dramatically when $K/N$ increases from $0$ to $0.2$, and then decays slowly between $0.2$ and $0.6$. When $K/N$ becomes approximately larger than $0.6$, the decaying speeds up again. The trend is clear: as more arms are allowed to be pulled at the same time, greedily selecting the best of them will make the performance more far away from the optimal.
    \item The PPP performs not quite well for very small $K$, but greatly improves as $K$ grows. This is consistent with our theoretical results in Section \ref{sec:offline}, where we show that the ratio of our designed PPP has a $\cO(1/\sqrt{K})$ gap with $1$. The shape of lines also have common features across different $N$'s. There is a small deterioration when $K$ grows from $1$ to $2$ and $3$. When $K$ approaches $4$ and $5$, PPP is comparable to the greedy policy. As $K$ grows larger, the performance consistently increases and approaches $1$. The trend is clear: as more arms are allowed to be pulled at the same time, smartly choosing a single period for each arm will make the performance near-optimal.
\end{enumerate}

We would like to provide some additional discussions on the performance when $K$ is small. In Section \ref{ssec:offline-statement}, we provide an example showing that the approximation ratio for the greedy policy may be arbitrarily close to $0$. While this holds true, it happens in the worst case. Our experiments suggest that when $N$ is large, $K$ is small, and the recovery mechanism is not that bizarre as in Example \ref{example:greedy:bad}, the arms may have enough time to recover from low rewards to higher ones. As a result, the greedy policy may still perform quite well or even near-optimal. On the contrary, the class of PPP may lack some flexibility when $K$ is small because the periods have to be taken as a specific form for small $a$. Nevertheless, our design of PPP owns the property of $1/2$ worst-case ratio, which does not hold for the greedy policy.

\subsection{Online Learning} \label{sec:simulation:online}

\subsubsection*{Settings}
We consider four different choices of $N\in\{50, 100, 150, 250\}$ and two different choices of $K\in\{5, 10, 25\}$. For each $(N, K)$, an instance $\cR$ is simulated using the same rules for the offline planning problem (see Section \ref{sec:simulation:offline}). For fixed $i$ and $d$, the \textit{random} reward for arm $i$ with idle time $d$ is sampled according to a triangular distribution with parameter $(0, 2R_i(d), R_i(d))$. More precisely, the probability density function is
\begin{align*}
    p(x) = \frac{\max\{\min\{x, 2R_i(d)-x\}, 0\}}{R_i^2(d)}, \quad \forall x\in\mathbb R.
\end{align*}

\subsubsection*{Algorithms} We consider the algorithm proposed in Section \ref{sec:online}. To improve the empirical performance, we make two amendments to Algorithm \ref{alg:online}. First, we always let $\epsilon=0$. That is, we always solve the program (\ref{program:knapsack}) to optimal. Second, we separately run $a=\lfloor\sqrt K\rfloor$ and $a=\lceil\sqrt K\rceil$ with $K'=K+1$ instead of using a single $a=\arg\max_a\frac{aK}{(a+1)(K+1)}$. Finally, we integrate Algorithm \ref{alg:online-1} into Algorithm \ref{alg:online}. That is, we also execute Line 7-8 in Algorithm \ref{alg:online-1} when determining the optimal schedule in each phase. This means in each phase we will output $5$ different $K$-PPP's and select the $K$-PPP that attains the largest long-run average reward among the $5$ results. The crucial tuning parameter $\phi$ is traversed within $\{50, 100, \cdots, 1000\}$.

\subsubsection*{Implementation Details}
We fix $T=10000$. For each $N$, we sample an instance $\cR$ through the procedure described above. The universal upper bound $R_{\max}$ is taken as 
\begin{align*}
    100\cdot\left\lceil\frac{\max_{i, d}R_i(d)}{50}\right\rceil,
\end{align*}
which is the smallest multiples of $100$ no less than the random reward for any arm $i$. Then for each $(K, \phi)$ pair, we run the algorithm for $50$ times to compute the long-run average reward. We compute the ratio between the average reward and the upper bound $\UB_\cR[K]$ and average them over $50$.

\subsubsection*{Results} 

\begin{figure}[!ht]
    \centering
    \includegraphics[width=16cm]{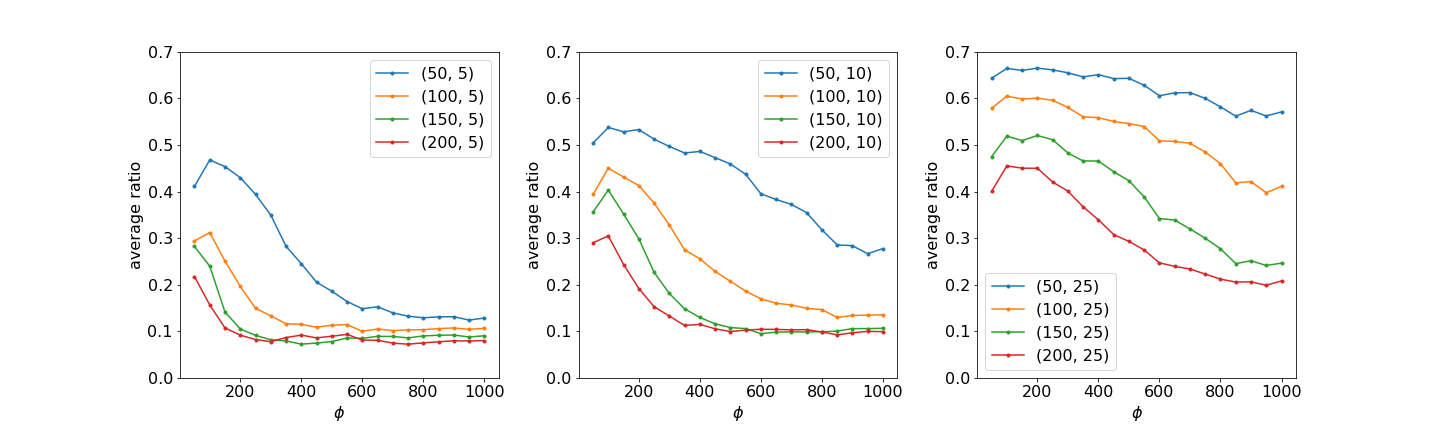}
    \caption{Online ratios}
    \label{fig:online}
\end{figure}

We plot the ratios in Figure \ref{fig:online} as $\phi$ changes for different pairs of $(N,K)$. In each subfigure of Figure \ref{fig:online}, $K$ is fixed while $N$ varies. There are $4$ lines in total. Each line corresponds to the ratio for fixed $(N, K)$ with various $\phi$. Some observations are as follows.
\begin{enumerate}
    \item The ratio for each fixed $(N, K)$ has a clear tendency: it first increases as $\phi$ increases, and then decreases as $\phi$ further increases. The optimal $\phi$ is approximately $100$ across all the scenarios, which is in accordance with our theoretical prediction that $\phi$ should be approximately $\sqrt T$. 
    \item The comparison between different $(N, K)$'s also show some illustrative phenomenon. For fixed $K$, as $N$ increases, the ratio decreases. In this case, the intuition is that learning among more arms becomes harder as $N$ increases. For fixed $N$, as $K$ increases, the ratio increases because of the contribution from better planning. This is consistent with our offline planning result that the (long-run) approximation ratio of PPP approaches $1$ as $K$ grows.
\end{enumerate}

We also hope to point out that the ratios are not satisfactory especially when $K$ is small (say, $K=5$) and $N$ is large (say, $N=200$). This can be caused by three reasons. First, when $K$ is small, the ratio gap between PPP and the upper bound is not negligible. Second, when $N$ is large compared to $K$, a learning algorithm has to take more steps for exploration to obtain valuable information about all arms. Third, in our algorithm design, the UCB bound is constructed by a universal upper bound $R_{\max}$ on any arm $i$ and any idle time $d$, which may cause the learning algorithm to be too conservative and explore too much. Though this can be alleviated by using arm-dependent bounds, such bounds may depend on accurate prior experience.

\section{Conclusion}

In this work, we consider the problem of dynamic planning and learning under a general bandit model of non-stationary recovering rewards. The non-stationarity stems from both the time elapsed and the policy itself. Solving the offline problem where all recovery functions are known is computationally hard, so we focus on a simple class of ``Purely Periodic Policies''. We develop a new framework for rounding and scheduling to obtain a policy with provable performance guarantee. The long-run approximation ratio is shown to be uniformly lower bounded by $1/2$ and asymptotically optimal with respect to the number of arms allowed to be pulled. We then show how to solve the online learning problem with $\widetilde\cO(\sqrt T)$ regret compared to the offline benchmark. We design our algorithm through a novel combination of our offline result, the knapsack problem, and upper confidence bounds, bypassing the difficulties of planning-while-learning under recovering rewards.

There is some interesting future work in line. For the offline planning problem, it is intriguing to see if we can build other classes of policies other than the class of PPP such that the approximation can be improved. What is the optimal approximation ratio we can achieve using polynomial-time algorithms? In the online learning problem, it may be worth investigating whether the regret bound can be improved to $\widetilde\cO(\sqrt{NKT})$. Particularly, the regret bound may be further improved if additional structures are added among different arms. We believe our work opens up new insights for future work to solve broader planning and learning problems with non-stationary and recovering rewards.

%
%
%

\bibliographystyle{informs2014} 
\bibliography{main}

\ECSwitch \small


\ECHead{Proofs of Statements}

\section{Proof of NP-Hardness of the offline problem}
Here we adopt the similar idea in \cite{cella2020stochastic_supp} to give a proof of NP-Hardness of the offline problem, even if we confine ourselves to find a long-run optimal policy within the class of purely periodic policies.

{\noindent\bf Proof.} 

Our proof relies on a reduction from the Periodic Maintenance
Scheduling Problem (PMSP) to our problem. In PMSP, we are given $n$ machines for service, and $n$ positive integers $\ell_1, \cdots, \ell_n$ such that $\sum_{i=1}^n1/\ell_i \leq 1$. We call $\{\ell_i\}_{i\in[N]}$ is feasible, if there exists a schedule such that the consecutive service times of each machine $i$ are exactly $\ell_i$ times apart, and meanwhile in each time period at most 1 machine is in service. The question is to examine whether $\{\ell_i\}_{i\in[N]}$ is feasible. \citet{bar2002minimizing} showed that PMSP is NP-complete.

Given an instance of PMSP with $\ell_1, \cdots, \ell_n$, we prove that $\{\ell_i\}_{i\in[N]}$ is feasible if and only if there exists a $1$-PPP such that its long-run average reward is $\sum_{i\in[N]}1/\ell_i$. We let $N=n$, $K=1$, and
\begin{align*}
    R_i(d) = \left\{
    \begin{matrix}
    0, & \text{if }d < \ell_i, \\
    1, & \text{if }d \geq \ell_i.
    \end{matrix}
    \right.
\end{align*}

On one hand, if this instance of PMSP is feasible, then we can directly apply the corresponding schedule to pull arms, yielding a long-run average reward
\begin{align*}
    \sum_{i\in[N]}R_i(\ell_i)/\ell_i = \sum_{i\in[N]}1/\ell_i.
\end{align*}
Moreover, this schedule is purely periodic.

On the other hand, suppose we can find a purely periodic schedule of arms pulling such that the long-run average reward is no less than $\sum_{i\in[N]}1/\ell_i$. Note that the long-run average reward of pulling an arm $i$ is upper bounded by
\begin{align*}
    R_i(d)/d \leq 1/\ell_i\ (\forall d\geq 1),
\end{align*}
and the equality holds iff $d = \ell_i$. Therefore, we must pull arm $i$ every $\ell_i$ times eventually, or the average reward within one period is strictly less than $1/\ell_i$. This means that the instance is feasible.

\section{Proofs of Lemmas and Theorems}




{\noindent\bf Proof of Lemma \ref{lemma:F-property}.}

Let $F_{i, T}^{\cc}(x)$ be the value of the following problem.
\begin{align} \label{program:upper-bound-single-cc}
     \max_s \quad & \frac{1}{T}\sum_{j=1}^J R_i^{\cc}(s_j-s_{j-1}) \\
    \text{s.t.} \quad & J \leq x\cdot T, \nonumber\\
    & 0 = s_0 < s_1 < \cdots < s_J \leq T. \nonumber
\end{align}
Intuitively, $F_{i, T}^\cc(x)$ is the optimal average reward of $i$ under $\{R_i^\cc(d)\}$, given that we pull arm $i$ no more than $x\cdot T$ times. Then we can see that $F_{i, T}^\cc(x)\geq F_{i, T}(x)\ (\forall i\in[N], T\geq 1, x\in[0, 1])$.

\noindent\textbf{Claim 1.} Fix $i\in[N]$, $T\geq 1$ and $x\in[0, 1]$, then there exists an optimal solution to (\ref{program:upper-bound-single-cc}) such that $s_J=T$ and $\{s_j-s_{j-1}\}$ can take at most two different values.

Apparently, under Assumption \ref{assumption:mono}, the objective value will never decrease if we let $s_J = T$. Thus, we can always assume $s_J=T$. For simplicity, we write $ds_j\triangleq s_j - s_{j-1}$. For any feasible solution $\{s_j\}$ and $J$ of (\ref{program:upper-bound-single}), we define 
\[
\text{dist}(\{s_j\}_{j=1}^J) = \sum_{1\leq i, j\leq J} \left|ds_j - ds_i\right|\cdot\mathds{1}\left\{\left|ds_j - ds_i\right| > 1\right\}.
\]
Let $\{s_j^*\}_{j=1}^{J^*}$ be an optimal solution such that $\text{dist}\left(\{s_j^*\}_{j=1}^{J^*}\right)$ attains the minimum among all optimal solutions. This is attainable since the number of feasible solutions to (\ref{program:upper-bound-single-cc}) is finite, and as a result, there always exists an optimal solution and the number of optimal solutions is finite. Suppose $\text{dist}\left(\{s_j^*\}_{j=1}^{J^*}\right) > 0$, then we choose $j_1=\arg\min_j\left\{ds_j^*\right\}$ and $j_2=\arg\max_j\left\{ds_j^*\right\}$. Without loss of generality, we assume $j_1 < j_2$. Then $ds_{j_2}^* - ds_{j_1}^* \geq 2$. We define a new set $\{s_j'\}$ as follows.
\begin{align*}
    s_j' = \left\{
    \begin{matrix}
    & s_j^* + 1, & \quad \text{if }j_1\leq j < j_2, \\
    & s_j^*, & \quad \text{else}.
    \end{matrix}
    \right.
\end{align*}
Note that $s_J' = s_J^* = T$. Then 
\begin{align*}
    ds_{j_1}' & = ds_{j_1}^* + 1 \leq ds_{j_2}^* - 1 = ds_{j_2}', \\
    ds_j' & = ds_j^*, \quad \forall j\neq j_1, j_2.
\end{align*}
By our choice of $j_1$ and $j_2$ ,we have
\begin{align*}
    \left|ds_{j_k}' - ds_{j}'\right|\mathds{1}\left\{\left|ds_{j_k}' - ds_{j}'\right| > 1\right\} & \leq \left|ds_{j_k}^* - ds_{j}^*\right|\mathds{1}\left\{\left|ds_{j_k}^* - ds_{j}^*\right| > 1\right\}, \quad \forall k\in\{1, 2\}, \\
    \left|ds_{j_2}' - ds_{j_1}'\right|\mathds{1}\left\{\left|ds_{j_2}' - ds_{j_1}'\right| > 1\right\} & < \left|ds_{j_2}^* - ds_{j_1}^*\right| = \left|ds_{j_2}^* - ds_{j_1}^*\right|\mathds{1}\left\{\left|ds_{j_2}^* - ds_{j_1}^*\right| > 1\right\}.
\end{align*}
Thus $\text{dist}\left(\{s_j'\}_{j=1}^J\right) < \text{dist}\left(\{s_j^*\}_{j=1}^J\right)$. However, since $\{R_i^\cc(d)\}$ is concave, we have
\begin{align*}
    & \quad R_i^\cc(ds_{j_1}') + R_i^\cc(ds_{j_2}') - R_i^\cc(ds_{j_1}^*) - R_i^\cc(ds_{j_2}^*) \\
    & = \left(R_i^\cc(ds_{j_1}') - R_i^\cc(ds_{j_1}'-1)\right) - \left(R_i^\cc(ds_{j_2}^*) - R_i^\cc(ds_{j_2}^*-1)\right) \geq 0.
\end{align*}
This means either $\{s_j^*\}_{j=1}^{J^*}$ is not optimal, or it does not have the minimum dist value. A contradiction. Therefore, $\text{dist}\left(\{s_j^*\}_{j=1}^{J^*}\right) = 0$, indicating $\{ds_j^*\}$ must take at most two different values.

\noindent\textbf{Claim 2.} Let $x = \alpha\frac{1}{d+1} + (1-\alpha)\frac{1}{d}$, where $d\in\mathbb Z_+, d\geq d_i^{(1)}$, and $\alpha\in(0, 1]$. Then $F_{i, T}^\cc(x)\leq \alpha\frac{R_i^\cc(d+1)}{d+1} + (1-\alpha)\frac{R_i^\cc(d)}{d} = F_i(x)$.

From Claim 1, $\exists d'\in\mathbb Z_+$, an optimal scheduling of (\ref{program:upper-bound-single}) satisfies $s_j-s_{j-1}\in\{d', d'+1\}$. Then we have
\begin{align*}
    T \leq (d'+1)J\leq (d'+1)xT < (d'+1)T/d,
\end{align*}
which indicates $d' \geq d$. Suppose in the optimal scheduling, 
\begin{align*}
    a & = \#\left\{j\in[J]: s_j-s_{j-1} = d'\right\}, \\
    b & = \#\left\{j\in[J]: s_j-s_{j-1} = d'+1\right\}.
\end{align*}
Then
\begin{align*}
    a + b \leq x\cdot T, \quad ad' + b(d'+1) = T.
\end{align*}
Since $\{R_i^\cc(d)\}_{d\geq 0}$ is concave, we have $R_i^\cc(d)/d$ is non-increasing. If $d'\geq d+1$, then we have
\begin{align*}
    F_{i, T}(x) & = \frac{a}{T}R_i^\cc(d') + \frac{b}{T}R_i^\cc(d'+1)  = \frac{ad'}{T}\frac{R_i^\cc(d')}{d'} + \frac{b(d'+1)}{T}\frac{R_i^\cc(d'+1)}{d'+1} \\
    & \leq \frac{R_i^\cc(d+1)}{d+1} \leq \alpha\frac{R_i^\cc(d+1)}{d+1} + (1-\alpha)\frac{R_i^\cc(d)}{d}.
\end{align*}
If $d' = d$, then
\begin{align*}
    T = ad + b(d+1) \leq (d+1)xT - a,
\end{align*}
which means $\frac{ad}{T}\leq 1-\alpha$. We thus have
\begin{align*}
    F_{i, T}(x) & = \frac{a}{T}R_i^\cc(d) + \frac{b}{T}R_i^\cc(d+1) = \frac{ad}{T}\frac{R_i^\cc(d)}{d} + \frac{b(d+1)}{T}\frac{R_i^\cc(d+1)}{d+1} \\
    & = \frac{ad}{T}\frac{R_i^\cc(d)}{d} + \left(1 - \frac{ad}{T}\right)\frac{R_i^\cc(d+1)}{d+1} \leq \alpha\frac{R_i^\cc(d+1)}{d+1} + (1-\alpha)\frac{R_i^\cc(d)}{d},
\end{align*}
since $\frac{R_i^\cc(d)}{d}\geq \frac{R_i^\cc(d+1)}{d+1}$.

We are left to show that $\alpha\frac{R_i^\cc(d+1)}{d+1} + (1-\alpha)\frac{R_i^\cc(d)}{d} = F_i(x)$, which means $\alpha\frac{R_i^\cc(d+1)}{d+1} + (1-\alpha)\frac{R_i^\cc(d)}{d}$ can be achieved in an asymptotic sense. Let $k\geq 1$ such that $d_i^{(k)}\leq d < d+1\leq d_i^{(k+1)}$. Then
\begin{align*}
    & \quad \alpha\frac{R_i^\cc(d+1)}{d+1} + (1-\alpha)\frac{R_i^\cc(d)}{d} \\
    & = \alpha\frac{R_i^\cc(d_i^{(k)})\frac{d_i^{(k+1)} - (d+1)}{d_i^{(k+1)}-d_i^{(k)}} + R_i^\cc(d_i^{(k+1)})\frac{(d+1) - d_i^{(k)}}{d_i^{(k+1)}-d_i^{(k)}}}{d+1} \\
    & \quad + (1-\alpha)\frac{R_i^\cc(d_i^{(k)})\frac{d_i^{(k+1)} - d}{d_i^{(k+1)}-d_i^{(k)}} + R_i^\cc(d_i^{(k+1)})\frac{d - d_i^{(k)}}{d_i^{(k+1)}-d_i^{(k)}}}{d} \\
    & = R_i^\cc(d_i^{(k)})\frac{xd_i^{(k+1)} - 1}{d_i^{(k+1)} - d_i^{(k)}} + R_i^\cc(d_i^{(k+1)})\frac{1 - xd_i^{(k)}}{d_i^{(k+1)} - d_i^{(k)}} \\
    & = R_i(d_i^{(k)})\frac{xd_i^{(k+1)} - 1}{d_i^{(k+1)} - d_i^{(k)}} + R_i(d_i^{(k+1)})\frac{1 - xd_i^{(k)}}{d_i^{(k+1)} - d_i^{(k)}}
\end{align*}

We solve
\begin{align*}
    a + b = x\cdot T, \quad ad_i^{(k)} + bd_i^{(k+1)} = T
\end{align*}
and get $a = \frac{d_i^{(k+1)}xT - T}{d_i^{(k+1)}-d_i^{(k)}}$ and $b = \frac{T - d_i^{(k)}xT}{d_i^{(k+1)}-d_i^{(k)}}$. Then
\begin{align*}
    \lfloor a\rfloor + \lfloor b\rfloor \leq x\cdot T, \quad \lfloor a\rfloor d_i^{(k)} + \lfloor b\rfloor d_i^{(k+1)} \leq T.
\end{align*}
We have
\begin{align*}
    \lim\inf_T F_{i, T}(x) & \geq \lim_T \frac{\lfloor a\rfloor R_i(d_i^{(k)})}{T} + \frac{\lfloor b\rfloor R_i(d_i^{(k+1)})}{T} \\
    & = R_i(d_i^{(k)})\frac{xd_i^{(k+1)} - 1}{d_i^{(k+1)} - d_i^{(k)}} + R_i(d_i^{(k+1)})\frac{1 - xd_i^{(k)}}{d_i^{(k+1)} - d_i^{(k)}} \\
    & = \alpha\frac{R_i^\cc(d+1)}{d+1} + (1-\alpha)\frac{R_i^\cc(d)}{d}.
\end{align*}

\noindent\textbf{Claim 3.} Let $x\geq \frac{1}{d_i^{(1)}}$. Then $F_{i, T}(x)\leq \frac{R_i(d_i^{(1)})}{d_i^{(1)}} = F_i(x)$.

We can see from the definition of $d_i^{(1)}$ that
\begin{align*}
    F_{i, T}(x) & \leq \frac{1}{T}\sum_{j=1}^J R_i(s_j-s_{j-1}) = \frac{1}{T}\sum_{j=1}^J \frac{R_i(s_j-s_{j-1})}{s_j-s_{j-1}}(s_j-s_{j-1}) \leq \frac{R_i(d_i^{(1)})}{d_i^{(1)}}.
\end{align*}
On the other hand, if we let $s_j = j\cdot d_i^{(1)}$, then 
\begin{align*}
\lim\inf_T F_{i, T}(x) \geq \lim_T \frac{\lfloor T/d_i^{(1)}\rfloor}{T}R_i(d_i^{(1)}) = \frac{R_i(d_i^{(1)})}{d_i^{(1)}}.
\end{align*}

$\hfill\Box$

{\noindent\bf Proof of Lemma \ref{lemma:upper-bound}.} 

We first prove the claim that (\ref{program:upper-bound}) is an upper bound on the original problem. For any schedule within a finite time horizon $T$, let 
\begin{align*}
    x_i = \#\{t\in[T]: i\text{ is pulled at time period } t\}/T\in[0, 1].
\end{align*}
Then $\sum_{i\in[N]}x_i\leq K$ always hold. Further, the reward collected from arm $i$ is no less than $F_{i, T}(x_i)\leq F_{i}(x_i)$, by our definition of $F_{i, T}(x_i)$ and Lemma \ref{lemma:F-property}. Thus, the total reward collected is no less than
\begin{align*}
    \sum_{i\in[N]}F_i(x_i).
\end{align*}


Now we prove the remaining part. Suppose $\{x_i^*\}$ is a feasible solution of (\ref{lemma:upper-bound}). Then making $x_i^*\gets \min\left\{x_i^*, 1/d_{i, 1}\right\}$ does not decrease the objective value. If more than one components are not of the form $\{1/d_i^{(k)}\}\cup\{0\}$, then we can assume $x_{i_1}^*\in(1/(d_{i_1, j_1+1}), 1/d_{i_1, j_1})$ and $x_{i_2}^*\in(1/(d_{i_2, j_2+1}), 1/d_{i_2, j_2})$, where $i_1\neq i_2$. From Lemma \ref{lemma:F-property}, $F_{i_k}\ (k\in\{1, 2\})$ is linear on $[1/(d_{i_k, j_k+1}), 1/d_{i_k, j_k}]$, so we can move $x_{i_1}$ larger (smaller) and $x_{i_2}$ smaller (larger) by the same distance until one of them reach an endpoint. The objective value will not decrease for at least one direction, and meanwhile this will not violate the hard constraint, but strictly decrease the number of $i\in[N]$ that $x_i^*\notin \{1/d_i^{(k)}\}\cup\{0\}$ in the feasible solution. We can thus repeat the procedure above until the solutions is transformed into the property stated in Lemma \ref{lemma:upper-bound}. In fact, the procedure takes at most $\cO(N)$ time to transform any feasible solution into the form we want.

$\hfill\Box$

{\noindent\bf Proof of Lemma \ref{lemma:rounding-lower-bound}.}


When $1/x_i^*\in\{d_i^{(k)}\}_{k\geq 1}$, we have
\begin{align*}
\frac{R_i(d_i)/d_i}{F_i(x_i^*)} = \frac{R_i(d_i)/d_i}{R_i^{\cc}(1/x_i^*)x_i^*} = \frac{R_i(d_i)/d_i}{R_i(1/x_i^*)x_i^*} \geq \frac{1}{d_ix_i^*},
\end{align*}
where the first equality holds from Lemma \ref{lemma:F-property}, and the inequality holds from Assumption \ref{assumption:mono}. We notice that
\begin{align*}
    \left\{1, \cdots, a-1\right\}\bigcup\left\{a\times 2^\ell, (a+1)\times 2^\ell, \cdots, (2a-1)\times 2^\ell\right\}_{\ell\geq 0}\subset \mathcal D[a]
\end{align*}
because any positive integer number no less than $2a$ can be written as a positive odd number (less than $2a$) times a power of 2. Now if $1/x_i^* \leq 2a-1$, then $d=1/x_i^*$. If $1/x_i^* \geq 2a$, then
\begin{align*}
    d_i\leq \sup_{a\leq b < 2a}\frac{b+1}{b}\cdot 1/x_i^* \leq \frac{a+1}{ax_i^*}.
\end{align*}
Thus, $\frac{1}{d_ix_i^*}\geq \frac{a}{a+1}$.

$\hfill\Box$

{\noindent\bf Proof of Lemma \ref{lemma:power-of-2}.}

\noindent\textbf{Part 1.} Without loss of generality, we assume that the sum of frequencies is strictly larger than 1 (otherwise we simply choose $\mathcal I_{j_1} = \mathcal I_j$). For each $1\leq k\leq |\mathcal I_j|$, we write $d_{i_k} = (2j-1)\times 2^{\ell_{i_k}}$. Since the sum of frequencies is no less than 1, there exists some $1\leq m < |\mathcal I_j|$ such that
\begin{align*}
    \sum_{k=1}^m 1/d_{i_k} \leq 1 < \sum_{k=1}^{m+1} 1/d_{i_k}.
\end{align*}
We will prove in the following that
\begin{align*}
    \sum_{k=1}^m 1/d_{i_k} = 1.
\end{align*}
In fact, we have
\begin{align*}
     d_{i_m} - 1 \leq d_{i_m} - d_{i_m}/d_{i_{m+1}} < \sum_{k=1}^m d_{i_m}/d_{i_k} = \sum_{k=1}^m2^{\ell_{i_m}-\ell_{i_k}}\in \mathbb Z_+,
\end{align*}
which indicates that
\begin{align*}
    d_{i_m} \leq \sum_{k=1}^m d_{i_m}/d_{i_k}.
\end{align*}
This is what we desire. Apparently, we can find $m$ by adding $d_{i_k}$ one by one and compare each sum with 1. Moreover, once the sum reaches 1 (our proof above guarantees this), we can make the former $m$ products into a group, and restart from product $i_{m+1}$. The total time complexity is $\mathcal O(|\mathcal I_j|)$.

\noindent{\bf Part 2.} We begin with $j=1$. We write $d_{i_k} = 2^{\ell_{i_k}}$ and let $\ell = \max_{i\in\mathcal I_1}\ell_i$. We use induction method to prove that after sorting, we can specify a 1-PPP in $\mathcal O(|\mathcal I_1|\log d_{i_{|\mathcal I_1|}})$ time. When $\ell=0$, there is only one product in $\mathcal I_1$ with frequency 1. Let $t_{i_1}=0$. The result is correct.

Suppose for $\ell$ the result is correct. Now consider the case for $\ell+1$. Then $\ell_{i_1}\geq 1$ and 
\begin{align*}
    \sum_{i\in\mathcal I_1}1/(d_i/2) \leq 2.
\end{align*}
If $\sum_{i\in\mathcal I_1}1/(d_i/2) \leq 1$, then by induction, we can specify a 1-PPP in $\mathcal O\left(|\mathcal I_1|\log \left(d_{i_{|\mathcal I_1|}}/2\right)\right)$ time. We project the offering time by $t\to 2t$. The total time complexity is 
\begin{align*}
    \mathcal O\left(|\mathcal I_{1}|\log \left(d_{i_{|\mathcal I_{1}|}}/2\right)\right) + \mathcal O\left(|\mathcal I_{1}|\right) = \mathcal O\left(|\mathcal I_{1}|\log d_{i_{|\mathcal I_{1}|}}\right).
\end{align*}
If $\sum_{i\in\mathcal I_1}1/(d_i/2) > 1$, then applying the proof of Part 1, we can split $\mathcal I_1$ into two parts $\mathcal I_{11}$ and $\mathcal I_{12}$ in $\mathcal O(|\mathcal I_1|)$ time such that
\begin{align*}
    \sum_{i\in\mathcal I_{1k}} 1/(d_i/2) \leq 1, \quad \forall k\in\{1, 2\}.
\end{align*}
By induction, we can specify a feasible 1-PPP for $\mathcal I_{11}$ and $\mathcal I_{12}$. The time complexity for this procedure is
\begin{align*}
    \mathcal O\left(|\mathcal I_{11}|\log \left(d_{i_{|\mathcal I_{1}|}}/2\right)\right) + \mathcal O\left(|\mathcal I_{12}|\log \left(d_{i_{|\mathcal I_{1}|}}/2\right)\right) = \mathcal O\left(|\mathcal I_{1}|\log \left(d_{i_{|\mathcal I_{1}|}}/2\right)\right).
\end{align*}
Now we project the offering time by $t\to 2t-1$ for products in $\mathcal I_{11}$ and $t\to 2t$ for products in $\mathcal I_{12}$. This fulfills our requirement. The total time complexity is 
\begin{align*}
    \mathcal O\left(|\mathcal I_{1}|\log \left(d_{i_{|\mathcal I_{1}|}}/2\right)\right) + \mathcal O\left(|\mathcal I_{1}|\right) = \mathcal O\left(|\mathcal I_{1}|\log d_{i_{|\mathcal I_{1}|}}\right).
\end{align*}

We continue on general cases. When $j>1$, we notice that
\begin{align*}
    \sum_{i\in\mathcal I_j}1/\left(d_i/(2j-1)\right) \leq 2j-1.
\end{align*}
By Part 1, we can split $\mathcal I_j$ into at most $2j-1$ disjoint sets  such that $\mathcal I_j = \bigcup_s \mathcal I_{j_s}$ such that
\begin{align*}
    \sum_{i\in\mathcal I_{j_s}} 1/\left(d_i/(2j-1)\right)\leq 1, \quad \forall s.
\end{align*}
By our proof for $j=1$ above, we can specify a feasible 1-PPP for $\mathcal I_{j_s}\ (\forall s)$. We project the offering time by $t\to (2j-1)(t-1)+s$ for products in $\mathcal I_{j_s}$. This completes the construction. The total time complexity is
\begin{align*}
    \mathcal O(|\mathcal I_j|) + \sum_{s}\mathcal O\left(|\mathcal I_{j_s}|\log \left(d_{i_{|\mathcal I_j|}}/(2j-1)\right)\right) = \mathcal O\left(|\mathcal I_j|\log d_{i_{|\mathcal I_j|}}\right).
\end{align*}

$\hfill\Box$

{\noindent\bf Proof of Theorem \ref{theorem:offline}.}

For each selected product $i\in[N]$, it is offered at time $t_i+kd_i\ (k\geq 1)$ until $T$. Thus, the number of time it is offered is lower bounded by
\begin{align*}
    \lfloor (T-t_i)/d_i\rfloor \geq \lfloor T/d_i\rfloor > T/d_i - 1.
\end{align*}
The total reward of $i$ throughout the whole time horizon is lower bounded by
\begin{align*}
    (T/d_i - 2)\cdot R(d_i) = R_i(d_i)/d_i\cdot T - 2\cdot R(d_i).
\end{align*}
Therefore, the total reward obtained is lower bounded by
\begin{align*}
    & \quad \sum_{i\text{ is selected}}F_i(1/d_i)\cdot T - 2\sum_{i\text{ is selected}}R(d_i) \\
    & = \frac{\sum_{i\text{ is selected}}F_i(1/d_i)}{\UB[N, K]}\cdot \UB[N, K]\cdot T - \cO(N) \\
    & \geq \gamma_K\cdot \UB[N, K]\cdot T - \cO(N).
\end{align*}

$\hfill\Box$

{\noindent\bf Proof of Theorem ``tightness''.}

Consider the following instance of blocking bandit problem consisting of two types of arms.
\begin{itemize}
    \item ``Small'' arms: $K'=K-1$ stationary arms with mean reward $R_i(d) = 1/\left(\sqrt{K\ln K}\right)\ (\forall i\in[K'], \forall d\geq 1)$. Here, ``stationary'' means the reward distribution remains the same irrespective of the time elapsed as well as the policy used.
    \item ``Big'' arms: $N' = \lfloor\sqrt{K/\ln K}\rfloor$ arms with mean reward $R_{K'+i}(d) = p_i\mathds 1\left\{d\geq p_i\right\}\ (\forall i\in[N'])$, where $N'\leq p_1 < \cdots < p_{N'}$ are $N'$ smallest prime numbers that are no less than $N'$.
\end{itemize}

\noindent \textbf{Claim 1.} $p_i\leq 2\sqrt{K\ln K}\ (\forall i\in[N'])$.

It suffices to prove that $p_{N'}\leq 3\sqrt{K\ln K}$. From our choice of $N'$, we can see that $p_{N'}$ is at most the $2N'$th smallest prime number. When $N'\geq 3$, $K > 3$. By Theorem 3 of \citet{rosser1962approximate}, we have
\begin{align*}
    p_{N'} & \leq 2N'\ln 2N' + 2N'\ln \ln \left(2N'\right) \\
    & \leq 2\sqrt{\frac{K}{\ln K}}\ln\sqrt{\frac{K}{\ln K}} + 2\sqrt{\frac{K}{\ln K}}\ln\ln\sqrt{2\frac{K}{\ln K}} \\
    & \leq \sqrt{K\ln K} - \sqrt{\frac{K}{\ln K}}\ln\ln K + 2\sqrt{\frac{K}{\ln K}}\ln\ln K \\
    & \leq 2\sqrt{K\ln K}.
\end{align*}
If $N'=1$, then
\begin{align*}
    p_{N'} = 2 \leq 2\sqrt{2\ln 2}\leq 2\sqrt{K\ln K}.
\end{align*}
If $N'=2$, then $\sqrt{K/\ln K}\geq 2$, yielding $K\geq 3$. Thus, 
\begin{align*}
    p_{N'} = 3 \leq 2\sqrt{3\ln 3}\leq 2\sqrt{K\ln K}.
\end{align*}

Now we turn to the optimal solution of the upper bound (\ref{program:upper-bound}). Notice that
\begin{align*}
    K'\times 1/1 + \sum_{i\in[N']}1/p_i \leq K - 1 + N'\cdot 1/N' = K.
\end{align*}
For each small arm $i\ (i\in[K-1])$, period $d_i^*=1$ (or frequency $1/d_i^*=1/1$) is optimal for the single arm. For each big arm $K'+i\ (i\in[N'])$, period $d_{K'+i} = p_i$ (or frequency $1/d_i^*=1/p_i$) is also optimal for the single arm. Thus, in the optimal solution, $x_i^* = 1/1\ (\forall i\in[K'])$ while $x_i^* = 1/p_{K'+i}\ (\forall i\in[N'])$. The long-run average reward is
\begin{align*}
    \UB[N, K] = \frac{K'}{\sqrt{K\ln K}} + N' \leq 2\sqrt{K/\ln K}.
\end{align*}

A $K$-PPP can be regarded as a projection as follows: for any arm $i$ with frequency $1/d_i^*$ $\forall i\in[K'+N']$, we project it with 3 types of operations: (i) rounding the $1/d_i^*$ to some $1/d_i$ where $d_i>d_i^*$; (ii) abandoning arm $i$, i.e., $d_i=+\infty$; (iii) maintaining the $1/d_i^*$, i.e., $1/d_i = 1/d_i^*$. Note that when we round $1/d_i^*$ to some $1/d_i$ where $d_i<d_i^*$, the resulting reward will be $0$, which can be regarded as abandoning arm $i$.

The procedure above can be further regarded as a two-stage process. We first do (ii) and (iii), and then for the remaining arms we do (i). Note that after each stage, the objective value is non-increasing. Suppose after the first stage, $x$ out of ``small'' arms remains and $y$ out of ``big'' remains. 

If $y < N'$, then the remaining objective value is no more than $\UB[N, K] - 1$, leading to the ratio no more than
\begin{align*}
    1 - \frac{1}{2\sqrt{K/\ln K}}.
\end{align*}
If $x < K - 1 - \sqrt{K/\ln K}/3$, then the remaining objective value is no more than
\begin{align*}
    \UB[N, K] - (K-1-x)\cdot\frac{1}{\sqrt{K\ln K}} \leq \UB[N, K] - \frac{1}{3\ln K}.
\end{align*}
As a result, the ratio is no more than
\begin{align*}
    1 - \frac{1}{6\sqrt{K\ln K}}.
\end{align*}

In the following, we assume $x\geq K-1-\sqrt{K/\ln K}/3$ and $y=N'$.

\noindent\textbf{Claim 2.} To obtain a $K$-PPP, we must do (i) on at least $x + y - K$ arms. 

In fact, let's assume that we do (i) on $x_1$ ``small'' arms and $y_1$ ``big'' arms, and then obtain a $K$-PPP, represented by $\{d_i\}$ and $\{t_i\}$. This means that the remaining arms that are neither abandoned nor operated still constitute a $K$-PPP. The number of these arms is $x-x_1+y-y_1$. Among these arms, each ``big ''arm $i$ has a period $d_i$ equals a prime number. The prime numbers are mutually different. Therefore, whatever the choice of $\{t_i\}$, there always exists some period $t$ such that all the $y-y_1$ ``big'' arms are pulled. At time $t$, $x-x_1$ ``small'' arms with period 1 are also pulled. As a result, at time $t$ at least $x-x_1+y-y_1$ arms are pulled simultaneously. This suggests that $x-x_1+y-y_1\leq K$, yielding
\begin{align*}
    x_1 + y_1\geq x + y - K.
\end{align*}

If we operate (i) on a ``small'' arm, the loss for this arm is lower bounded by
\begin{align*}
    1/\sqrt{K\ln K}\cdot (1/1-1/2) = 1/\left(2\sqrt{K\ln K}\right).
\end{align*}
If we operate (i) on a ``big'' arm, the loss for this arm is lower bounded by
\begin{align*}
    p_i\cdot (1/p_i - 1/(p_i+1)) \geq 1/\left(2\sqrt{K \ln K} + 1\right).
\end{align*}
Thus, the overall long-run loss is lower bounded by
\begin{align*}
    (x + y - K)/\left(2\sqrt{K\ln K} + 1\right) & \geq \left(2\sqrt{K/\ln K}/3 - 2\right)/\left(2\sqrt{K \ln K} + 1\right).
\end{align*}
The approximation ratio is then upper bounded by
\begin{align*}
    & \quad 1 - \frac{\left(2\sqrt{K/\ln K}/3 - 2\right)/\left(2\sqrt{K \ln K} + 1\right)}{2\sqrt{K/\ln K}} \\
    & \leq 1 - \frac{1}{6\sqrt{K\ln K}+3} + \frac{1}{2K}.
\end{align*}
$\hfill\Box$

{\noindent\bf Proof of Lemma \ref{lemma:knapsack}.}
 Consider the following more general problem. 
\begin{align} \label{program:knapsack-general}
    \max_x \quad & \sum_{i\in[N]}\sum_{s\in\mathcal S} r_{i, s}x_{i, s} \\
    \text{s.t.} \quad & \sum_{i\in[N]}\sum_{s\in\mathcal S}w_{i, s}x_{i, s} \leq K', \nonumber\\
    & \sum_{s\in\mathcal S}x_{i, s} \leq 1, \quad \forall i\in[N], \nonumber\\
    & x_{i, s}\in\{0, 1\}, \quad \forall i\in[N], \forall s\in\mathcal S. \nonumber
\end{align}
We first assume that $r_{i, s}\in\mathbb Z_+\cup\{0\}$, then we let $v(n, r)$ be the value of following problem.
\begin{align*}
    \min_x \quad & \sum_{i\in[n]}\sum_{s\in\mathcal S} w_{i, s}x_{i, s} \\
    \text{s.t.} \quad & \sum_{i\in[n]}\sum_{s\in\mathcal S}r_{i, s}x_{i, s} = r, \\
    & \sum_{s\in\mathcal S}x_{i, s} \leq 1, \quad \forall i\in[n], \\
    & x_{i, s}\in\{0, 1\}, \quad \forall i\in[n], \forall s\in\mathcal S.
\end{align*}
If the problem is infeasible, we let $v(n, r) = +\infty$, then we have the following recurrence formula:
\begin{align*}
    v(n, r) = \min_{s: r_{n, s}\leq r}\left\{v(n-1, r), w_{n, s} + v(n-1, r - r_{n, s})\right\}.
\end{align*}
We also have the initial conditions:
\begin{align*}
    v(1, r) & = \left\{
    \begin{array}{ll}
    w_{1, s}, & \quad \text{if }r = r_{1, s}, \\
    +\infty, & \quad \text{else},
    \end{array}
    \right. \\
    v(n, 0) & = 0, \quad\forall n\in[N].
\end{align*}
Let $r_{\max} = \max_{i, s}r_{i, s}$, then the largest possible $r$ is upper bounded by $Nr_{\max}$. Thus, $v(n, r)$ can be computed within $\cO\left(N^2r_{\max}|\mathcal S|\right)$ time. The maximal reward for (\ref{program:knapsack-general}) is then computed by iterating through $\{v(n, r)\}_{n\in[N], r\leq Nr_{\max}}$ such that $v(n, r)\leq K'$ while $r$ is maximized. Finding the optimal solution requires tracing back $v(n, r)$ to the initial conditions, which consumes $\cO(N|\mathcal S|)$ time. Thus, the total time complexity is $\cO\left(N^2r_{\max}|\mathcal S|\right)$.

For the general case, we define $\tilde r_{i, s} = \lfloor\frac{Nr_{i, s}}{\epsilon r_{\max}}\rfloor$. We compute $(\ref{program:knapsack-general})$ with $\{r_{i, s}\}$ replaced by $\{\tilde r_{i, s}\}$. Then since
\begin{align*}
    0 \leq\frac{\frac{N}{\epsilon r_{\max}}r_{i, s} - \tilde r_{i, s}}{\frac{N}{\epsilon r_{\max}}r_{i, s}} \leq \frac{1}{\frac{Nr_{i, s}}{\epsilon r_{\max}}} \leq \epsilon, \quad \forall i\in[N], s\in\mathcal S,
\end{align*}
The solution we compute is a $(1-\epsilon)$-optimal solution of (\ref{program:knapsack-general}). As a final step, we substitute $\mathcal S$ with $\mathcal D_\phi[a]$, and the computation time is
\begin{align*}
    \cO\left(N^2\max_{i, s}\left\lfloor\frac{Nr_{i, s}}{\epsilon r_{\max}}\right\rfloor|\mathcal D_\phi[a]|\right) = \cO\left(\frac{N^3a\log_2\phi}{\epsilon}\right).
\end{align*}

$\hfill\Box$

{\noindent\bf Proof of Theorem \ref{theorem:online}.}

For completeness, we restate some definitions. Let $n_{i, j}(d)$ be the number of samples we have collected for $R_i(d)$ from the beginning of the whole time horizon to the end of phase $j$. Here we let $n_{i, 0}(d) = 0$ for all $i\in[N]$ and $d\in\mathbb Z_+$. At the beginning, we have a natural upper bound $R_i(d)\leq R_{\max}$. Let
\begin{align*}
    \bar R_{i, j-1}(d) \triangleq  \frac{\sum_{\ell=1}^{n_{i, j-1}(d)}\hat R_i^\ell(d)}{n_{i, j-1}(d)}
\end{align*}
be the empirical mean of $R_i(d)$ calculated by the samples collected prior to phase $j$. Here, $\hat R_i^\ell(d)$ is the $\ell$th sampled reward we collected when we offer product $i$ $d$ time periods after we offered it last time. The upper bound $\hat R_{i, j}(d)$ is then computed by
\begin{align*}
    \min\left\{\bar R_{i, j-1}(d) + R_{\max}\sqrt{\frac{2\log (KT)}{\max\left\{n_{i, j-1}(d), 1\right\}}}, \ R_{\max}\right\}.
\end{align*}

Let $\mathcal G$ be the ``good event'' that $\forall i\in[N]$, all phases $j$ and all $d\in\mathcal D_\phi[a]$, the following holds:
\begin{align} \label{event-good}
    \left|R_i(d) - \bar R_{i, j-1}(d)\right| \leq R_{\max}\sqrt{\frac{2\log (KT)}{\max\left\{n_{i, j-1}(d), 1\right\}}}.
\end{align}
Since $\forall i\in[N]$, after each phase, we update the estimation of at most one element in $\{R_i(d)\}_{d\in\mathcal D_\phi[a]}$, and so by Hoeffding's inequality,
\begin{align*}
    \mathbb P\left(\mathcal G^c\right) & \leq N\cdot \left\lceil\frac{T}{\phi}\right\rceil\cdot 2\exp\left(-2\cdot 2\log (KT)\right) \\
    & \leq N\cdot \frac{2T}{\phi}\cdot 2\exp\left(-4\log (KT)\right) \leq \frac{4N}{\phi KT^3}.
\end{align*}
Thus, the total loss incurred when $\mathcal G^c$ occurs is bounded by
\begin{align*}
    \frac{4N}{\phi KT^3}R_{\max}KT = \cO\left(\frac{NR_{\max}}{\phi T^2}\right).
\end{align*}

Next, we consider the situation when $\mathcal G$ holds. Then we have
\begin{align*}
    \hat R_{i, j}(d) \geq R_i(d) \geq \hat R_{i, j}(d) - 2R_{\max}\sqrt{\frac{2\log (KT)}{\max\left\{n_{i, j-1}(d), 1\right\}}}
\end{align*}
because of (\ref{event-good}) and $R_i(d)\leq R_{\max}$. For brevity, we write $\gamma_\epsilon = \min\left\{1, \frac{K}{K'+a-1}\right\}(1-\epsilon)$. Denote $\hat\cR_j = \{\hat R_{i, j}(d)\}_{i, d}$. Then for each phase $j$, we have
\begin{align}
    & \quad \sum_{i\in[N]}\hat R_{i, j}(d_{i, j})/d_{i, j} \nonumber\\ 
    & \geq \sum_{i\in[N]}\hat R_{i, j}(d_{i, j}^\epsilon)/d_{i, j}^\epsilon\cdot\min\left\{1, \frac{K}{K'+a-1}\right\} \nonumber\\
    & \geq \min\left\{1, \frac{K}{K'+a-1}\right\}(1-\epsilon)\val_{\hat\cR_j}[K'|\cD_\phi[a]] \nonumber\\
    & \geq \gamma_{\epsilon}\val_\cR[K'|\cD_\phi[a]]. \label{formula:phase-lower-bound-1}
\end{align}
The first inequality holds because of the following reason. For all $i$, $d_{i, j}^\epsilon$ already belongs to $\cD[a]$. Since the sum of the frequencies $\{d_{i, j}^\epsilon\}$ is no more than $K'$, the resulting number of $1$-PPPs in the R-S procedure is upper bounded by $K'+a-1$. Then we are selecting $K$ out of at most $K'+a-1$ $1$-PPPs that attain the largest long-run average reward. The second inequality holds because we are solving (\ref{program:knapsack}) up to an $\epsilon$ precision. The last inequality holds because $\hat\cR_j\geq \cR$ on $\cD_\phi[a]$.

The reward obtained at a given phase $j$ is
\begin{align*}
    & \quad \phi\sum_{i=1}^NR_i(d_{i, j})/d_{i, j} - \cO(NR_{\max}) \\
    & = \phi\gamma_{\epsilon}\val_\cR[K'|\cD_\phi[a]] - \phi\left(\gamma_{\epsilon}\val_\cR[K'|\cD_\phi[a]] - \sum_{i=1}^N R_i(d_{i, j})/d_{i, j}\right) - \cO(NR_{\max}) \\
    & \geq \phi\gamma_{\epsilon}\val_\cR[K'|\cD_\phi[a]] - \phi\sum_{i=1}^N\left(\hat R_{i, j}(d_{i, j})/d_{i, j} - R_i(d_{i, j})/d_{i, j}\right) - \cO(NR_{\max}) \\
    & \geq \phi\gamma_{\epsilon}\val_\cR[K'|\cD_\phi[a]] - \phi\sum_{i=1}^N\cO\left(\frac{R_{\max}\sqrt{\log (KT)}}{d_{i, j}\sqrt{\max\left\{n_{i, j-1}(d_{i, j}), 1\right\}}}\right) - \cO(NR_{\max}),
\end{align*}
where the first inequality is from (\ref{formula:phase-lower-bound-1}). Summing over all phases, we can derive that the reward obtained under $\mathcal G$ is lower bounded by
\begin{align*}
    & \quad T\gamma_{\epsilon}\val_{\cR}[K'|\cD_\phi[a]] - \sum_{i=1}^N \sum_{j}\cO\left(\frac{\phi R_{\max}\sqrt{\log (KT)}}{d_{i, j}\sqrt{\max\left\{n_{i, j-1}(d_{i, j}), 1\right\}}}\right) - \cO\left(\frac{NR_{\max}T}{\phi}\right) \\
    & = T\gamma_{\epsilon}\val_{\cR}[K'|\cD_\phi[a]] - \sum_{i=1}^N \sum_{d\in\mathcal D_\phi[a]}\sum_{j: d_{i, j} = d}\cO\left(\frac{\phi R_{\max}\sqrt{\log (KT)}}{d_{i, j}\sqrt{\max\left\{n_{i, j-1}(d_{i, j}), 1\right\}}}\right) - \cO\left(\frac{NR_{\max}T}{\phi}\right)
\end{align*}
Note that for all $j$ such that $d_{i, j}=d$, we can list in the increasing order $\{j_0, j_1, j_2, \cdots\}\ (j_0=0)$, and we have
\begin{align*}
    n_{i, j_\ell-1}(d) \geq n_{i, j_{\ell-1}}(d) \geq n_{i, j_{\ell-1}-1}(d) + \left\lceil\frac{\phi}{d}\right\rceil - 1 \geq n_{i, j_{\ell-1}-1}(d) + \frac{\phi}{3d} \geq \frac{\phi}{3d}, \quad \forall \ell\geq 2.
\end{align*}
Thus, we have
\begin{align*}
    \frac{\phi}{d\sqrt{\max\{n_{i, j_2-1}(d), 1\}}} \leq \frac{\phi}{d\sqrt{n_{i, j_2-1}(d)}} \leq 3\sqrt{n_{i, j_2-1}(d)},
\end{align*}
and $\forall \ell\geq 3$,
\begin{align*}
    \frac{\phi}{d\sqrt{\max\{n_{i, j_{\ell}-1}(d), 1\}}} \leq 3\frac{n_{i, j_\ell-1}(d) - n_{i, j_{\ell-1}-1}(d)}{\sqrt{n_{i, j_\ell-1}(d)}} \leq 6 \frac{n_{i, j_\ell-1}(d) - n_{i, j_{\ell-1}-1}(d)}{\sqrt{n_{i, j_\ell-1}(d)} + \sqrt{n_{i, j_{\ell-1}-1}(d)}}
\end{align*}
Then we have
\begin{align*}
    & \quad\sum_{j: d_{i, j} = d}\cO\left(\frac{\phi}{d_{i, j}\sqrt{\max\left\{n_{i, j-1}(d_{i, j}), 1\right\}}}\right) \\
    & = \sum_{\ell}\cO\left(\frac{\phi}{d\sqrt{\max\left\{n_{i, j_\ell-1}(d), 1\right\}}}\right) \\
    & = \sum_{\ell=1}\cO\left(\frac{\phi}{d\sqrt{\max\left\{n_{i, j_\ell-1}(d), 1\right\}}}\right) + \sum_{\ell=2}\cO\left(\frac{\phi}{d\sqrt{\max\left\{n_{i, j_\ell-1}(d), 1\right\}}}\right) + \sum_{\ell\geq 3}\cO\left(\frac{\phi}{d\sqrt{\max\left\{n_{i, j_\ell-1}(d), 1\right\}}}\right) \\
    & \leq \cO\left(\frac{\phi}{d}\right) + \cO\left(\sqrt{n_{i, j_2-1}(d)}\right) + \sum_{\ell\geq 3}\cO\left(\frac{n_{i, j_\ell-1}(d) - n_{i, j_{\ell-1}-1}(d)}{\sqrt{n_{i, j_\ell-1}(d)} + \sqrt{n_{i, j_{\ell-1}-1}(d)}}\right) \\
    & \leq \cO\left(\frac{\phi}{d}\right) + \cO\left(\sqrt{n_i(d)}\right),
\end{align*}
where $n_i(d)$ is the number of samples we collected for $R_i(d)$ throughout the whole time horizon. Combined with the loss incurred under $\mathcal G^c$, the overall reward can be further bounded by
\begin{align*}
    & \quad T\gamma_{\epsilon}\val_{\cR}[K'|\cD_\phi[a]] - \sum_{i=1}^N \sum_{d\in\mathcal D_\phi[a]}\sum_{j: d_{i, j} = d}\cO\left(\frac{\phi R_{\max}\sqrt{\log (KT)}}{d_{i, j}\sqrt{\max\left\{n_{i, j-1}(d_{i, j}), 1\right\}}}\right) - \cO\left(\frac{NR_{\max}T}{\phi}\right) \\
    & \geq T\gamma_{\epsilon}\val_{\cR}[K'|\cD_\phi[a]] - R_{\max}\sum_{i=1}^N \sum_{d\in\mathcal D_\phi[a]}\cO\left(\frac{\phi\sqrt{\log (KT)}}{d} + \sqrt{n_i(d)\log (KT)}\right) - \cO\left(\frac{NR_{\max}T}{\phi}\right) \\
    & \geq T\gamma_{\epsilon}\val_{\cR}[K'|\cD_\phi[a]] - R_{\max}\cO\left(\phi N\log (a+1)\sqrt{\log (KT)} + \sqrt{aKNT\log\phi\log (KT)} + \frac{NT}{\phi}\right),
\end{align*}
where in the last inequality we use
\begin{align*}
    \sum_{d\in\mathcal D_\phi[a]}\frac{1}{d} \leq \sum_{a'=1}^{a}\sum_{d\in\mathcal D_{a}}\frac{1}{d} \leq \sum_{a'=1}^{a}\frac{2}{a'} = \cO\left(\log (a+1)\right),
\end{align*}
and
\begin{align*}
    \sum_{i=1}^N\sum_{d\in\mathcal D_\phi[a]}\sqrt{n_i(d)} & \leq \sqrt{\sum_{i=1}^N\sum_{d\in\mathcal D_\phi[a]}n_i(d).} \sqrt{\sum_{i=1}^N\sum_{d\in\mathcal D_\phi[a]} 1} \\
    & \leq \sqrt{KT\cdot N\left|\mathcal D_\phi[a]\right|} = \cO\left(\sqrt{KNTa\log \phi}\right).
\end{align*}

$\hfill\Box$

{\noindent\bf Proof of Lemma \ref{lemma:online-offline-lower}.}

Let $\{x_i^*\}$ be an optimal solution of (\ref{program:upper-bound}). From Lemma \ref{lemma:upper-bound}, we can assume that at most 1 of its non-zero components $x_{i_0}^*$ is not in $\{1/d_i^{(k)}\}_{k\geq 1}$. We round $x_{i_0}^*$ to $\widetilde x_{i_0}^* = \min\left\{y\geq x_{i_0}^*: 1/y\in\{d_i^{(k)}\}_{k\geq 1}\right\}$. We apply Step 1 in Section \ref{ssec:offline-schedule} to $\{x_i^*\}_{i\neq i_0}\cup\{\widetilde x_{i_0}^*\}$ and obtain $\{1/d_i\}$ such that $d_i\in\mathcal D[a]$. Define $\{x_{i, j, d}\}$ as follows,
\begin{align*}
    x_{i, j, d} = \mathds{1}\left\{d = d_i\right\}.
\end{align*}
Then $\{x_{i, j, d}\}$ satisfies the constraints of (\ref{program:knapsack}), which means it is a feasible solution. Thus, the optimal objective value of (\ref{program:knapsack}) is lower bounded by
\begin{align*}
    & \quad \sum_{i\in[N]}\sum_{d\in\mathcal D_\phi[a]}\hat R_{i, j}(d)x_{i, j, d}/d \\
    & \geq \sum_{i\in[N], d_i\in\mathcal D_\phi[a]}R_i(d_i)/d_i \\
    & \geq \sum_{i\in[N]}R_i(d_i)/d_i - \sum_{i\in[N], d_i\notin\mathcal D_\phi[a]}R_i(d_i)/d_i \\
    & \geq \sum_{i\in[N], i\neq i_0}\frac{a}{a+1}F_i(x_i^*) + \frac{a}{a+1}F_i(\widetilde x_{i_0}^*) - N\frac{R_{\max}}{\phi/2} \\
    & \geq \sum_{i\in[N]}\frac{a}{a+1}F_i(x_i^*) - \frac{2NR_{\max}}{\phi} \\
    & = \frac{a}{a+1}\UB[N, K] - \frac{2NR_{\max}}{\phi},
\end{align*}
where the third inequality follows from Lemma \ref{lemma:rounding-lower-bound}.

$\hfill\Box$

{\noindent\bf Proof of Lemma \ref{lemma:bound-of-method}.}

Without loss of generality, we assume that $x_i^* > 0$ for all $i$. Otherwise, we can just ignore those $i$ with $x_i=0$.

We first prove the first inequality. Let 
\begin{align*}
    a^* = \arg\min_{a\in\{1, 2, 3\}}\{d: d\geq d_1^{(k_1)}, d=(2a-1)\times 2^\ell, \ell\geq 0\}
\end{align*}
Then from the proof of Lemma \ref{lemma:rounding-lower-bound}, we know that under $a=a^*$, the corresponding $d_1$ satisfies
\begin{align*}
    d_1^{(k)} \leq d_1 \leq \frac{3}{4} d_1^{(k_1)}.
\end{align*}
As a result, we have
\begin{align*}
    \max_{a}\cR(\cI_{a, 1}^*) \geq \cR(\cI_{a^*, 1}^*) \stackrel{\text{(a)}}{\geq} R_1(d_1)/d_1 \geq \frac{3}{4}R_1(d_1^{(k)}) / d_1^{(k)}.
\end{align*}
We give some explanations on (a). If arm $1$ is included in the selected $K$-PPP $\cI_{a^*, 1}^*$, then clearly the inequality holds. Otherwise, arm $1$ is not included in $\cI_{a^*, 1}^*$. By our algorithm design, under $a=a^*$, when constructing the $K$-PPP, we select $K$ PPP's that attains the largest long-run average reward. There must exist a $1$-PPP included in $\cI_{a^*, 1}^*$, and meanwhile has a long-run average reward no less than $R_1(d_1)/d_1$. Thus, we have shown that the second inequality always holds.

We then prove the two remaining inequalities using a unified view. Note that for $m=2\text{ or }3$, there will be no scheduling error because we always have
\begin{align*}
    \sum_i 1/d_i \leq \sum_{i} x_i \leq \sum_{i} x_i^* \leq 1.
\end{align*}
We divide $\{1\}\cup[2, +\infty)$ into $A_1$ and $A_2$, where
\begin{align*}
    A_1 & = \{1\}\bigcup_{\ell\geq 1}(3\times 2^{\ell-1}, 2^{\ell+1}], \\
    A_2 & = \{2\}\bigcup_{\ell\geq 1}(2^\ell, 3\times 2^{\ell-1}].
\end{align*}

Let $m\in\{2, 3\}$ and assume that when $m=2$, $x_1^*\leq 1/2$. Then we always have $1/x_i\in\{1\}\cup[2, +\infty)=A_1\cup A_2$ for any $i$. We will prove that
\begin{align*}
    \max_a\cR(\cI^*_{a, m}) \geq \frac{3}{5}\left(\sum_i R_i(1/x_i)x_i\right).
\end{align*}
Here, we abuse some notations since $x_1$ is in fact dependent on the method.

When $a=1$, we have
\begin{align*}
    \cR(\cI_{1, m}^*) & = \sum_i R_i(d_i)/d_i \\
    & = \sum_{i:1/x_i\in A_1} R_i(d_i)/d_i + \sum_{i:1/x_i\in A_2} R_i(d_i)/d_i \\
    & \geq \sum_{i:1/x_i\in A_1} R_i(1/x_i)/d_i + \sum_{i:1/x_i\in A_2} R_i(1/x_i)/d_i \\
    & \geq \frac{3}{4}\sum_{i:1/x_i\in A_1} R_i(1/x_i)x_i + \frac{1}{2}\sum_{i:1/x_i\in A_2} R_i(1/x_i)x_i.
\end{align*}
The first inequality holds because $d_i\geq 1/x_i$ for any $i$. The last inequality holds because when $1/x_i\in A_1$, $d_i\leq \frac{4}{3x_i}$; when $1/x_i\in A_2$, $d_i\leq \frac{2}{x_i}$. Similarly, when $a=2$, we have
\begin{align*}
    \cR(\cI_{2, 3}^*) \geq \frac{1}{2}\sum_{i:1/x_i\in A_1} R_i(1/x_i)x_i + \frac{2}{3}\sum_{i:1/x_i\in A_2} R_i(1/x_i)x_i.
\end{align*}
This is because when $1/x_i\in A_1$, $d_i\leq \frac{2}{x_i}$; when $1/x_i\in A_2$, $d_i\leq \frac{3}{2x_i}$. Therefore, we have
\begin{align*}
    & \quad \max_{a}\cR(\cI_{a, 3}^*) \\
    & \geq \frac{2}{5}\cR(\cI_{1, 3}^*) + \frac{3}{5}\cR(\cI_{2, 3}^*) \\
    & \geq \frac{2}{5} \left(\frac{3}{4}\sum_{i:1/x_i\in A_1} R_i(1/x_i)x_i + \frac{1}{2}\sum_{i:1/x_i\in A_2} R_i(1/x_i)x_i\right) + \frac{3}{5}\left( \frac{1}{2}\sum_{i:1/x_i\in A_1} R_i(1/x_i)x_i + \frac{2}{3}\sum_{i:1/x_i\in A_2} R_i(1/x_i)x_i\right) \\
    & \geq \frac{3}{5}\left(\sum_i R_i(1/x_i)x_i\right).
\end{align*}

Now if $m = 3$, we have
\begin{align*}
    \max_a \cR(\cI_{a, 3}^*) & \geq \frac{3}{5}\left(\sum_i R_i(1/x_i)x_i\right) \\
    & = \frac{3}{5}\left(R_1(1/x_1)x_1 + \sum_{i>1} F_i(x_i^*)\right) \\
    & = \frac{3}{5}\left( R_1(d_1^{(k_1+1)})/d_1^{(k_1+1)} + \sum_{i>1} F_i(x_i^*)\right)
\end{align*}
If $m=2$ and $x_1^*\leq 1/2$, we have
\begin{align*}
    \max_a \cR(\cI_{a, 2}^*) & \geq \frac{3}{5}\left(\sum_i R_i(1/x_i)x_i\right) \\
    & = \frac{3}{5}\left(R_1(1/x_1)x_1 + \sum_{i>1} F_i(x_i^*)\right) \\
    & \geq \frac{3}{5}\left( R_1(d_1^{(k_1)})x_1 + \sum_{i>1} F_i(x_i^*)\right) \\
    & = \frac{3}{5}\left(R_1(d_1^{(k_1)})x_1^* + \sum_{i>1} F_i(x_i^*)\right)
\end{align*}

$\hfill\Box$

{\noindent \bf Proof of Proposition \ref{prop:offline}.} We first assume that $x_1^* > 1/2$. In this case, $d_1^{(k_1)} = 1 < 1/x_1^* < 2$. Using $a=1$ and $m = 2$, we can see that $d_1 = 2$. Furthermore, 
\begin{align*}
    1/x_i^* = 1/x_i \leq d_i < 2/x_i = 2/x_i^*
\end{align*}
for any $i > 1$. Therefore, 
\begin{align*}
    \cR(\cI^*_{1, 2}) & = \sum_{i} R_i(d_i)/d_i \\
    & = R_1(d_1)/d_1 + \sum_{i>1} R_i(d_i)/d_i \\
    & = R_1(2)/2 + \sum_{i>1} R_i(d_i)/d_i \\
    & \geq (R_1(1)/1)/2 + \sum_{i>1} R_i(1/x_i^*)x_i^*/2 \\
    & = F_1(1/d_1^{(k)})/2 + \sum_{i>1}F_i(x_i^*)/2 \\
    & \geq \frac{1}{2}\left(\sum_i F_i(x_i^*)\right)
\end{align*}

In the following, we assume that $x_1^*\leq 1/2$. By Lemma \ref{lemma:bound-of-method}, we only need to prove the following:
\begin{align}
    & \max\left\{
    \frac{3}{4}R_1(d_1^{(k_1)})/d_1^{(k_1)}, \quad\frac{3}{5}\left(R_1(d_1^{(k_1)})x_1^* + \sum_{i>1}F_i(x_i^*)\right), \right.\nonumber\\
    & \left.\indent\indent\ \ \frac{3}{5}\left(R_1(d_1^{(k_1+1)})/d_1^{(k_1+1)} + \sum_{i>1}F_i(x_i^*)\right)
    \right\} \geq \frac{1}{2}\left(\sum_{i\in[N]}F_i(x_i^*)\right). \label{formula:2/5}
\end{align}
Note that we have
\begin{align*}
x_1^* & = \alpha\frac{1}{d_1^{(k_1)}} + (1-\alpha)\frac{1}{d_1^{(k_1+1)}}. \\
F_1(x_1^*) & = \alpha\frac{R_1(d_1^{(k_1)})}{d_1^{(k_1)}} + (1-\alpha)\frac{R_1(d_1^{(k_1+1)})}{d_1^{(k_1+1)}}.
\end{align*}
Let $u = R_1(d_1^{(k_1)})/d_1^{(k_1)}$, $v = R_1(d_1^{(k_1+1)})/d_1^{(k_1+1)}$,$w = \sum_{i>1}F_i(x_i^*)$. Then the left-hand side of (\ref{formula:2/5}) can be written as
\begin{align*}
    & \quad \max\left\{\frac{3}{4}u, \frac{3}{5}\left(\alpha u + (1-\alpha)R_1(d_1^{(k_1)})/d_1^{(k_1+1)} + w\right), \frac{3}{5}\left(v + w\right)\right\} \\
    & \geq \max\left\{\frac{3}{4}u, \frac{3}{5}\left(\alpha u + w\right), \frac{3}{5}\left(v + w\right)\right\} \\
    & \geq \frac{1}{\frac{4}{3}(\alpha-\alpha^2) + \frac{5}{3}\alpha + \frac{5}{3}(1-\alpha)}\left((\alpha-\alpha^2)u + \alpha(\alpha u + w) + (1-\alpha)v+w\right) \\
    & \geq \frac{1}{\frac{4}{3}(\alpha-\alpha^2) + \frac{5}{3}}\left(\alpha u + (1-\alpha)v + w\right) \\
    & \geq \frac{1}{2}\left(\sum_{i\in[N]}F_i(x_i^*)\right).
\end{align*}

$\hfill\Box$

{\noindent\bf Proof of Lemma \ref{lemma:knapsack-1}.} Let $d_{\max} = \max\{d:d\in\cD_{a, \phi}\}$. We re-write (\ref{program:knapsack}) as (\ref{program:knapsack-1-1})
\begin{align} \label{program:knapsack-1-1}
    \max_x \quad & \sum_{i\in[N]}\sum_{d\in\mathcal D_\phi[1]}\hat R_{i, j}(d)x_{i, j, d}/d \\
    \text{s.t.} \quad & \sum_{i\in[N]}\sum_{d\in\mathcal D_\phi[1]}\frac{d_{\max}}{d}x_{i, j, d}\leq d_{\max}, \nonumber\\
    & \sum_{d\in\mathcal D_\phi[1]}x_{i, j, d}\leq 1, \quad \forall i\in[N], \nonumber\\
    & x_{i, j, d}\in\{0, 1\}, \quad \forall i\in[N], d\in\mathcal D_\phi[1]. \nonumber
\end{align}
Note that $\frac{d_{\max}}{d}\in\mathbb Z_+$. Let $u(N, d_{\max})$ denote the objective value of (\ref{program:knapsack-1-1}), then $u(N, d_{\max})$ can be recursively computed as:
\begin{align*}
    u(n, d) = \max\left\{u(n-1, d), \max_{\substack{d'\in\cD_{a, \phi}\\d'\leq\frac{d_{\max}}{d}}}\left\{u\left(n-1, d - \frac{d_{\max}}{d'}\right) + \frac{\hat R_{i, j}(d')}{d'}\right\}\right\}, \quad \forall 1\leq n\leq N, 1\leq d\leq d_{\max}.
\end{align*}
Also, the boundary conditions can be written as:
\begin{align*}
    u(0, d) = u(n, 0) = 0, \quad \forall d\geq 1, n\geq 1.
\end{align*}

Thus, $u(n, d)$ can be computed within $\cO\left(Nd_{\max}|\cD_{a, \phi}|\right)$ time. Finding the optimal solution requires tracing back $u(n, d)$ to the initial conditions, which consumes $\cO(N|\cD_{a, \phi}|)$ time. Thus, the total time complexity is $\cO\left(Nd_{\max}|\cD_{a, \phi}|\right) = \cO\left(N\phi\log_2\phi\right)$.

$\hfill\Box$

{\noindent\bf Proof of Lemma \ref{lemma:online-offline-lower-1}.} 

Let $\{x_i^*\}$ be an optimal solution of (\ref{program:upper-bound}). We apply Algorithm \ref{alg:offline-1} to $\{x_i^*\}$ and obtain $\{1/d_i\}$. Then there exists $1\leq a^*\leq 3$ such that $d_i\in\mathcal D_{a^*}$. Define $\{x_{i, j, d}\}$ as follows,
\begin{align*}
    x_{i, j, d} = \mathds{1}\left\{d = d_i\right\}.
\end{align*}
Then $\{x_{i, j, d}\}$ satisfies the constraints of (\ref{program:knapsack}). Moreover, $d_i \geq 1/x_i^* \geq d_i^{(1)}\ (\forall i\in[N])$. Then 
\begin{align*}
    & \quad \max_{1\leq a\leq 3}\ \val_{\cR}[K|\cD_{a, \phi}] \\
    & \geq \val_{\cR}[K|\cD_{a^*, \phi}] \\
    & \geq \sum_{i\in[N]}\sum_{d\in\mathcal D_{a^*, \phi}}\hat R_{i, j}(d)x_{i, j, d}/d \\
    & = \sum_{i\in[N], d_i\in\mathcal D_{a^*, \phi}}R_i(d_i)/d_i \\
    & \geq \sum_{i\in[N]}R_i(d_i)/d_i - \sum_{i\in[N], d_i\notin\mathcal D_\phi[a]}R_i(d_i)/d_i \\
    & \geq \sum_{i\in[N]}\frac{1}{2}F_i(x_i^*) - N\frac{R_{\max}}{\phi/2} \\
    & = \frac{1}{2}\UB[N, K'] - \frac{2NR_{\max}}{\phi}.
\end{align*}

$\hfill\Box$

{\noindent \bf Proof of Proposition \ref{prop:online}.} The proof generally follows those in Theorem \ref{theorem:online}. To simplify the explanation, we only point out the modifications here.

The first modification is in (\ref{formula:phase-lower-bound-1}). Instead, we should let $\hat R_j = \{R_{i, j}(d)\}_{i\in[N], j\in\bigcup_{1\leq a\leq 3}D_{\phi, a}}$ and write
\begin{align*}
    \sum_{i\in[N]}\hat R_{i, j}(d_{i, j})/d_{i, j} \geq \sum_{i\in[N]} \max_{1\leq a\leq 3}\val_{\hat\cR_j}[K|\cD_{\phi, a}] \geq \max_{1\leq a\leq 3}\val_{\cR}[K|\cD_{\phi, a}].
\end{align*}
The first inequality follows from Line 8 in Algorithm \ref{alg:offline-1}. Note that $1/x_i\in\cD_{a^*, \phi}$ and $\sum_{i}x_i \leq K$, then from Lemma \ref{lemma:power-of-2}, there will be no rounding error or scheduling error when we execute Line 9. Also, we are solving the program to optimal. The second inequality holds because $\hat\cR_j\geq\cR$.

The second modification is that in all the formulas following (\ref{formula:phase-lower-bound-1}), we should replace $\gamma_{\epsilon}\val_\cR[K'|\cD_{\phi}[a]]$ with
\begin{align*}
    \max_{1\leq a\leq 3} \val_\cR[K|\cD_{\phi, a}].
\end{align*}

The third modification is that in all the formulas following (\ref{formula:phase-lower-bound-1}), when we encounter summing over $d\in\cD_\phi[a]$, we should replace it with
\begin{align*}
    d\in\cD_\phi[3] = \bigcup_{1\leq a\leq 3}\cD_{a, \phi}.
\end{align*}

With the modifications above, the expected overall reward of Algorithm \ref{alg:online-1} can be bounded as
\begin{align*}
    T\max_{1\leq a\leq 3}\val_{\cR}[K|\cD_{a, \phi}] - R_{\max}\cO\left(\phi N\log 4\sqrt{\log (KT)} + \sqrt{3KNT\log\phi\log (KT)} + \frac{NT}{\phi}\right)
\end{align*}
Using Lemma \ref{lemma:online-offline-lower-1} and let $\phi = \Theta\left(\sqrt{\frac{T}{\log(1+K)}}\right)$, the overall reward can be lower bounded as
\begin{align*}
    \frac{1}{2}\UB_\cR[K]\cdot T - \widetilde \cO(N\sqrt T).
\end{align*}

\end{document}